\documentclass[sn-mathphys,Numbered]{sn-jnl}

\usepackage{amsmath,amssymb,amsfonts}%
\usepackage{amsthm}%
\usepackage{mathrsfs}%
\usepackage[title]{appendix}%
\usepackage{textcomp}%
\usepackage{manyfoot}%
\usepackage{booktabs}%
\usepackage{algorithm}%
\usepackage{algorithmicx}%
\usepackage{algpseudocode}%
\usepackage{listings}%

\usepackage{booktabs} 
\usepackage{graphicx}
\usepackage{xspace}
\usepackage[normalem]{ulem}
\useunder{\uline}{\ul}{}
\usepackage{multirow}
\usepackage{lscape}
\usepackage{rotating}
\usepackage{supertabular}
\usepackage{longtable}
\usepackage{enumerate}
\usepackage{comment}
\usepackage{ragged2e}
\usepackage{array}
\PassOptionsToPackage{hyphens}{url}
\usepackage{url}
\usepackage{hyperref}
\usepackage{array}
\usepackage{ragged2e}
\usepackage{caption}
\newcolumntype{P}[1]{>{\RaggedRight\hspace{0pt}}p{#1}}
\usepackage{fontawesome}
\usepackage[table,xcdraw]{xcolor}

\usepackage{tcolorbox}
\usepackage[switch]{lineno}
\usepackage{xcolor,colortbl}

\newif\ifdraft
\drafttrue

\theoremstyle{thmstyleone}%
\theoremstyle{thmstyletwo}%

\theoremstyle{thmstylethree}%

\begin{document}
\ifdraft
  \newcommand{\didar}[1]{{\color{red}\emph{Didar: #1}}\xspace}
  \newcommand{\muneera}[1]{{\color{blue}\emph{Muneera: #1}}\xspace}
  \newcommand{\fra}[1]{{\color{magenta}\emph{Fra: #1}}\xspace}
  \newcommand{\rifat}[1]{{\color{cyan}\emph{Rifat: #1}}\xspace}
\else
  \usepackage[disable]{todonotes}
  \newcommand{\didar}[1]{}
  \newcommand{\muneera}[1]{}
  \newcommand{\fra}[1]{}
  \newcommand{\rifat}[1]{}
\fi

\title[Article Title]{Challenges and Solutions in AI for All}

    
\author*[1]{\fnm{Rifat Ara} \sur{Shams}}\email{rifat.shams@csiro.au}

\author[1]{\fnm{Didar} \sur{Zowghi}}\email{didar.zowghi@csiro.au}

\author[1]{\fnm{Muneera} \sur{Bano}}\email{muneera.bano@csiro.au}

\affil[1]{\orgname{CSIRO's Data61}, \orgaddress{\country{Australia}}}


\abstract{Artificial Intelligence (AI)'s pervasive presence and variety necessitate diversity and inclusivity (D\&I) principles in its design for fairness, trust, and transparency. Yet, these considerations are often overlooked, leading to issues of bias, discrimination, and perceived untrustworthiness. In response, we conducted a Systematic Review to unearth challenges and solutions relating to D\&I in AI. Our rigorous search yielded 48 research articles published between 2017 and 2022. Open coding of these papers revealed 55 unique challenges and 33 solutions for D\&I in AI, as well as 24 unique challenges and 23 solutions for enhancing such practices using AI. This study, by offering a deeper understanding of these issues, will enlighten researchers and practitioners seeking to integrate these principles into future AI systems.}

\keywords{Diversity, Inclusion, Artificial Intelligence, Systematic Literature Review}



\maketitle

\section{Introduction}
\label{sec:intro}

Artificial Intelligence (AI) has become a critical part of our society, presenting unique advantages and challenges. The ethical implications of AI, including fairness, trust, bias, and transparency are pressing issues that must be addressed. Research has indicated that AI systems can entrench and even exacerbate existing biases in areas such as criminal justice and recruitment processes \cite{bellamy2018ai, dattner2019legal}. Maintaining trust in AI is crucial for ensuring its widespread adoption, but the blackbox nature of these systems can undermine trust \cite{schmidt2020transparency, von2021transparency}. In response to these challenges, calls have been made for the deployment of ``fairness-aware'' algorithms that take demographic diversity into account and increase transparency in decision-making processes \cite{selbst2019fairness}.

The integration of diversity and inclusion (D\&I) principles in AI has the potential to mitigate the challenges posed by the lack of fairness and bias \cite{zowghi2023diversity}. Research suggests that diverse teams increase the likelihood of recognizing and addressing biases in AI systems \cite{bellamy2018ai}. From a design perspective, diverse teams bring different perspectives on fairness and can identify additional sources of bias in data or algorithms \cite{selbst2019fairness}. From a user's standpoint, involving marginalized communities in AI development can increase the likelihood of the technology being fair and trustworthy for those groups and increase its acceptance among them \cite{srinivasan2021biases}. Furthermore, ethical concerns for AI technology should also extend beyond privacy and transparency issues to include diversity and inclusion \cite{saheb2023ethically}.

The topics of bias and fairness in AI have received significant attention in recent years. Mehrabi et al. \cite{mehrabi2021survey} conducted a literature review on the sources of data and algorithm biases in AI applications and the different fairness definitions used to reduce bias in AI. Bertrand et al. \cite{bertrand2022cognitive} conducted an SLR on 37 papers exploring cognitive biases in Explainable AI (XAI) systems, and identified four ways cognitive biases impact XAI-assisted decisions. Another study \cite{xivuri2021systematic} reviewed 47 articles on fairness in AI algorithms and found a lack of consensus on definitions of AI algorithmic fairness. Obermeyer et al. \cite{obermeyer2019dissecting} addressed racial biases through algorithms by providing health and cost predictions for both Black and white patients. Benthall et al. \cite{benthall2019racial} proposed a method for group fairness interventions using unsupervised learning to mitigate racialized social inequality, social segregation, and stratification in machine learning.  

In contrast, limited research can be found that has explored the principles of diversity and inclusion (D\&I) in AI or by AI. To the best of our knowledge no systematic literature review has been conducted on this topic. In this paper, therefore, we fill the above-mentioned research gap and present a systematic literature review that provides the state of the art on the topic of AI and diversity and inclusion. Our aim is to explore challenges and solutions (guidelines/ strategies/ approaches/ practices) in the research literature focused on diversity and inclusion in AI (D\&I in AI) as well as the applications of AI for diversity and inclusion practices (AI for D\&I). To differentiate ``D\&I in AI'' and ``AI for D\&I'' while extracting challenges and solutions, we followed two different definitions of these two terms. For ``D\&I in AI'', we followed the definition provided by Zowghi and da Rimini \cite{zowghi2023diversity}: ``inclusion of humans with diverse attributes and perspectives in the data, process, system, and governance of the AI ecosystem''. On the other hand, we defined ``AI for D\&I'' as ``the applications of AI systems to enhance the diversity and inclusion practices in environment''.


The main contributions of this SLR include a rigorous search, selection and analysis of 48 articles published in the last six years (2017-2022) on the topic of D\&I in AI as well as AI for D\&I. We believe that the results of our exploration presented in this paper contribute to a deeper understanding of diversity and inclusion considerations in AI system development and deployment. Our findings from this SLR present:

\begin{itemize}
    \item 55 unique challenges and 33 unique solutions about D\&I in AI as well as 24 unique challenges and 23 unique solutions about the applications of AI for D\&I practices.
    \item The number of studies on AI for D\&I are significantly less than the number of studies on D\&I in AI. Moreover, not all papers that state challenges also propose solution for each challenge.
    \item `Gender' is the prominent diversity attribute in AI, whereas other attributes (e.g., race, ethnicity, language) are given less attention.
    \item `Health' is the most discussed domain in the literature, whereas other domains such as law, banking, and transportation are ignored in the literature.
    \item `Facial analysis' and `natural language processing' are the most discussed types of AI systems to address D\&I; other AI systems are ignored such as voice recognition and large language models.
    \item `Governance' related challenges and solutions are less discussed both for D\&I in AI and AI for D\&I.
\end{itemize}



\textbf{Paper Organization.} Section \ref{sec:background} describes the background of this research and the related work. Section \ref{sec:methodology} briefly explains our research method and Section \ref{sec:results} reports the findings of this study. We discuss the findings in Section \ref{sec:discussion}. Section \ref{sec:ttv} discusses the possible threats to validity of this research. Finally, the research is concluded with possible future research directions in Section \ref{sec:conclusions}.
\section{Background and Related Work}
\label{sec:background}


AI has emerged as a technological force that is continuously evolving and reshaping various societal structures \cite{pereira2023systematic}. In recent years, there's been a heightened focus on the importance of D\&I in AI \cite{avellan2020ai}, but the literature reveals that D\&I concerns are not consistently addressed in AI projects due to the lack of operational tools, and ambiguity around responsibilities in the AI development process \cite{stoyanovich2018follow}. Neglecting D\&I can have serious repercussions including harm to users and slowing AI adoption. Therefore, it's crucial for project teams and stakeholders to understand the criticality of D\&I in AI. The awareness of D\&I in AI will enable them to identify, monitor, and mitigate potential risks and challenges, thereby fostering an AI-literate society that can make informed decisions about the use and participation in AI systems across various contexts.

As the body of AI literature continues to expand, a growing number of traditional and systematic reviews reflects an increased focus on issues related to bias \cite{shrestha2022exploring, varsha2023can}, fairness \cite{richardson2021framework, xivuri2021systematic}, transparency \cite{laato2022explain}, and explainability \cite{anjomshoae2019explainable}. This focus arises from the acknowledgment that AI systems have the potential to reproduce and even exacerbate existing societal biases, leading to practices that can be unfairly discriminatory \cite{fosch2022diversity, nadeem2022gender}. Bias in AI systems has roots in numerous factors, most notably the utilization of datasets that lack comprehensive representation of the entire society, leading to outcomes that are skewed \cite{fosch2022diversity}. Additionally, the homogeneity of AI's development community, primarily being Western and male, can unintentionally inject biases into the design and programming of AI systems \cite{nadeem2022gender}. Addressing this imbalance, there is a growing recognition of diversity and inclusion as critical elements in AI development that can significantly contribute to mitigating these biases \cite{zowghi2023diversity}.

Despite the acknowledged importance of diversity and inclusion, there is a gap in the literature regarding how these principles can be practically implemented in AI systems. Fosch-Villaronga and Poulsen \cite{fosch2022diversity} define \textbf{D\&I in AI} as a \textit{multi-faceted concept that addresses both the technical and socio-cultural aspects of AI}. They highlight \textbf{diversity} as the representation of individuals with respect to socio-political power differentials such as gender and race. \textbf{Inclusion}, they suggest, is the representation of an individual user within a set of instances, with better alignment between a user and the options relevant to them indicating greater inclusion. This concept is further analyzed at three levels: the \textit{technical}, the \textit{community}, and the \textit{user}. The technical level considers if algorithms account for all necessary variables and if they classify users in a discriminatory manner. The community level examines diversity and inclusivity in AI development teams, looking at gender representation and diversity of backgrounds. Finally, the user level focuses on the intended users of the system and how the research and implementation process takes into account the stakeholders and their feedback, emphasizing the principles of Responsible Research and Innovation.

Zowghi and da Rimini \cite{zowghi2023diversity} provide a more detailed and normative definition of D\&I within the context of AI and present a set of guidelines for ensuring these principles are incorporated into the AI development process. The authors focus on a socio-technological perspective, recognizing that addressing issues of bias and unfairness requires a holistic approach that considers cultural dynamics and norms and involves end users and other stakeholders. They defined \textbf{D\&I in AI} as \textit{`inclusion' of humans with `diverse' attributes and perspectives in the data, process, system, and governance of the AI ecosystem}. \textbf{Diversity} refers to the representation of the differences in attributes of humans in a group or society. \textbf{Attributes} are known facets of diversity including (but not limited to) the protected attributes in Article 26 of the International Covenant on Civil and Political Rights (ICCPR), as well as race, color, sex, language, religion, national or social origin, property, birth or other status, and inter-sections of these attributes. \textbf{Inclusion} is the process of proactively involving and representing the most relevant humans with diverse attributes; those who are impacted by, and have an impact on, the AI ecosystem context.

According to Zowghi and da Rimini \cite{zowghi2023diversity}, diversity and inclusion in AI can be structured and conceptualized involving five pillars: humans, data, process, system, and governance. The humans pillar focuses on the importance of including individuals with diverse attributes in all stages of AI development. The data pillar highlights the need to be mindful of potential biases in data collection and use. The process pillar emphasizes the need for diversity and inclusion considerations during the development, deployment, and evolution of AI systems. The system pillar recognizes the necessity for the AI system to be tested and monitored to ensure it does not promote non-inclusive behaviors. The governance pillar underlines the importance of structures and processes that ensure AI development is compliant with ethical principles, laws, and regulations.

There is limited literature on how AI can help in enhancing D\&I \cite{mathis2021novel, chauhan2022role, buyl2022tackling, borgs2019algorithmic}, but there is no comprehensive definition in literature to present the concept. D\&I in AI, and AI for D\&I, create a synergistic cycle of progress that enriches both fields and their potential to effect meaningful change. AI, functioning as a mirror, reflects the patterns and prejudices ingrained in our societies, revealing biases that often go unnoticed. This heightened visibility aids in improving D\&I by identifying gaps, promoting awareness, and guiding mitigation strategies. On the flip side, the integration of D\&I within AI's development process is equally critical. A diverse team of creators and evaluators can identify, understand, and correct underlying biases, resulting in more equitable and inclusive AI systems. Thus, D\&I and AI form a continuous, self-enhancing cycle: the use of AI advances D\&I, while fostering D\&I within AI development ensures more holistic, fair, and representative AI systems.

Even with these insights, many existing AI ethics guidelines remain narrowly focused on fairness, justice, and non-discrimination, with a heavy lean towards compliance-based procedures \cite{cachat2023diversity}. Furthermore, there is an evident gap in initiatives that aim to directly impact AI actors' behaviors and foster diversity, equity, and inclusion (DEI) awareness \cite{robert2020designing}. In terms of inclusivity, it is pertinent to note that the global discourse on AI often lacks voices and perspectives from the Global South and other underrepresented groups, with a marked dominance of Western perspectives \cite{roche2022ethics}. This imbalance affects the development of ethical AI standards and calls for more inclusive practices and deeper consideration of power structures in AI policy formulation \cite{nyariro2022integrating, mhlambi2023decolonizing, ormond2023governance}.

Despite the increased awareness of these concerns, there remains a dearth of comprehensive understanding in current research addressing these critical areas. Hence, the urgent need for a systematic literature review that investigates diversity and inclusion in AI. This approach will provide a comprehensive evaluation and synthesis of all existing research on this topic, which traditional literature reviews may fail to capture in their entirety. Consequently, it will help identify the current state of the art, define challenges and solutions, and shape future research directions, thereby addressing this critical gap in the literature.

\section{Methodology}
\label{sec:methodology}

This study aims to explore and gain a comprehensive understanding of diversity and inclusion in the context of artificial intelligence and the use of artificial intelligence for diversity and inclusion from the published research literature. Our research was guided by the following two research questions. \\

   \indent \textit{\textbf{RQ1.} What challenges and solutions are found in the literature about diversity and inclusion in AI (D\&I in AI)?}\\
   \indent \textit{\textbf{RQ2.} What challenges and solutions are found in the literature about the applications of AI for diversity and inclusion practices (AI for D\&I)?}\\

       \begin{figure*}[!htbp]
            \centering
            \includegraphics[width=0.85\textwidth]{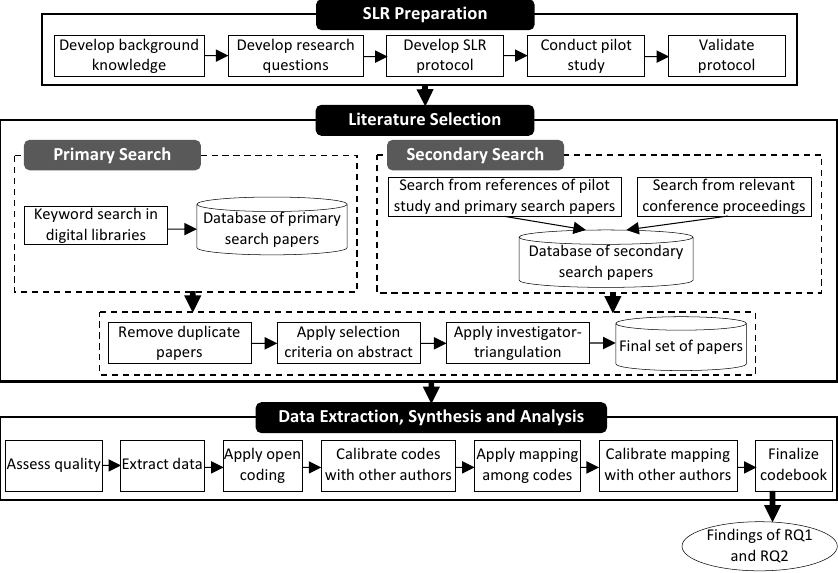}
            \caption{An overview of the research method}
            \label{fig:method}
    \end{figure*}

\begin{figure*}[!htbp]
        \centering
        \includegraphics[width=1\textwidth]{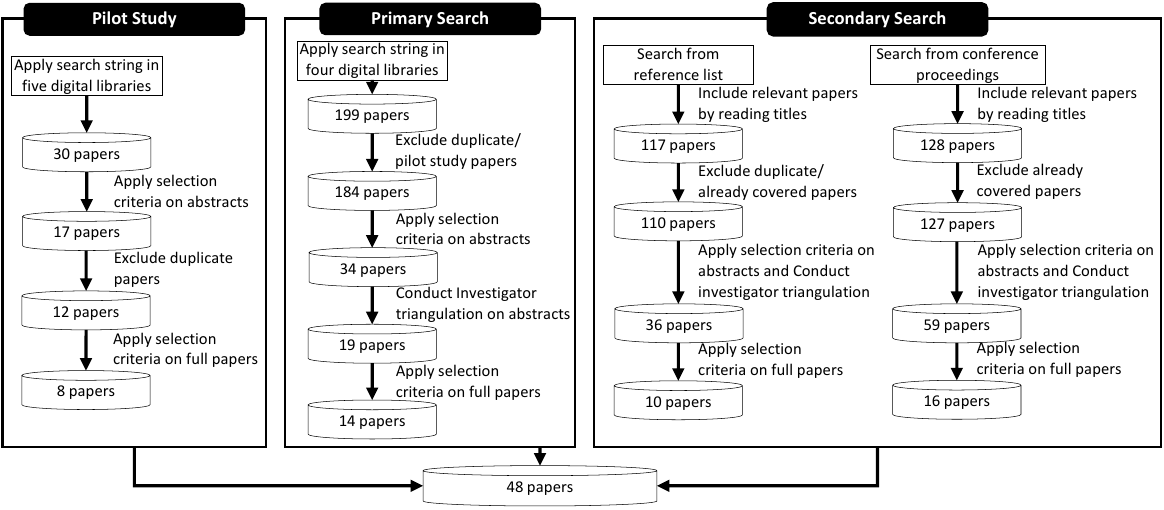}
        \caption{SLR paper selection summary}
        \label{fig:total_papers}
    \end{figure*}

We conducted a Systematic Literature Review in accordance with the guidelines established by Kitchenham et al. \cite{kitchenham2007guidelines} to address the research questions. This approach was chosen to comprehensively identify, evaluate, and interpret existing research in this under-explored area \cite{kitchenham2007guidelines}. The methodology of the SLR is outlined in \autoref{fig:method}. The preparation stage of the SLR involved the development of a background understanding of diversity and inclusion (D\&I) in AI, the formulation of research questions, the creation of an SLR protocol, and the validation of the protocol through a pilot study. The paper selection summary for the pilot study and the main study (primary search and secondary search \cite{jalali2012systematic}) is shown in \autoref{fig:total_papers}. As a result of a rigorous search and selection process, we finally identified 48 papers that satisfied inclusion/exclusion criteria and are relevant to D\&I in AI or AI for D\&I.

To ensure the validity of the data extraction process and the relevance of the search keywords, a pilot study was conducted at the outset of the process. The search string in the five digital libraries (ACM Digital Library, IEEE Xplore, Science Direct, Scopus, and Google Scholar) was formulated using the three primary keywords relevant to the research questions: ``artificial intelligence'', ``machine learning'', ``diversity and inclusion'' (see \autoref{appendix_B}). In order to guide the selection of studies, clear inclusion and exclusion criteria were established. The inclusion criteria were: ``papers on diversity and inclusion in AI or AI for diversity and inclusion'', ``papers in the form of peer reviewed published scientific papers (journal/conference)'', and ``papers published in 2017-2022''. The exclusion criteria were: ``papers not related to diversity and inclusion in AI or AI for diversity and inclusion'', ``literature review paper'', ``Tutorial/workshop paper/ArXiv paper/magazine article/book/book chapter'', ``Master/Ph.D. dissertations'', ``conference version of a study that has an extended journal version'', ``papers not written in English'', ``full papers unavailable online'', and ``papers already covered in the pilot study''. We included papers from 2017-2022, as we did not find many relevant papers prior to 2017. Moreover, we identified only one relevant paper in 2017, whereas the majority of the studies published in 2022.

Considerations for D\&I in AI or AI for D\&I as a stand-alone topic of research are scarce in the literature. We experimented with including the terms ``bias'' and/or ``fairness'' in our search string which resulted in a very large number of papers. For example, ACM digital library returned 92 research articles on diversity and inclusion in AI (2017-2022). When we added ``bias'' or ``fairness'' to the search string, it returned 669305 articles\footnote{Search date: 29/08/2022}. To ensure the feasibility of the SLR, we decided to narrow the scope and remove ``bias'' and ``fairness'' from the search string. 


\subsection{Primary Search}
In the pilot study, we used the keyword ``diversity and inclusion'' which was restricted to the papers that were based on both ``diversity'' and ``inclusion''. After several rounds of discussion among the authors, we decided to include all the papers on ``diversity'' OR ``inclusion'', so that no paper was left out which worked on either diversity or inclusion in AI. Therefore, we developed the final search string for our main study using the three main keywords (``diversity'', ``inclusion'' and ``artificial intelligence'') and their corresponding alternatives. For example, we used ``machine learning'' as an alternative to ``artificial intelligence''. Similarly, we used two alternatives of the keyword, ``inclusion'': ``inclusive'' and ``inclusiveness''. The primary search was carried out with this search string in four digital libraries: ACM Digital Library, IEEE Xplore, Science Direct, and Scopus. We also applied the same search string in Google Scholar, but it provided the papers which were already covered in the above-mentioned four digital libraries. The search string was customized depending on the interfaces of different digital libraries. The details of the primary search protocol and the search results are shown in \autoref{appendix_B}.

After eliminating duplicates in the primary search, a total of 184 papers underwent a rigorous application of the study selection criteria on the abstracts, resulting in a selection of 34 relevant papers (\autoref{fig:total_papers}). The next stage of selection process was guided by the principle of investigator triangulation \cite{archibald2016investigator}, where all the authors read the 34 abstracts independently and made decisions on inclusion/exclusion. Finally, they discussed their opinions and agreed on the final selection of 19 papers which later underwent a selection process by reading the full papers. Then, the first author carefully evaluated the full text of each of the included studies and excluded 5 papers, as they were found to be irrelevant to the research questions (such as diverse literature, diverse algorithms, diverse technology). Finally, 14 papers were selected from the primary search for data extraction.
    

\subsection{Secondary Search}
The secondary searches involved a manual examination of the titles of the references listed in the selected pilot and primary studies. In addition, a manual scan was performed on the proceedings of two most frequent conferences where the pilot and primary studies were published: ACM Conference on Fairness, Accountability, and Transparency and AAAI/ACM Conference on AI, Ethics, and Society. After removing the duplicate papers and the papers already covered in the pilot study and primary search, we came up with a total of 237 papers (110 from the reference list and 127 from the conference proceedings). Then, study selection criteria were applied to the abstracts to yield 95 papers from the secondary search. Investigator triangulation was also met to validate our selection. Then, the first author evaluated the full text of each of the included studies and excluded 69 papers from the secondary search due to their irrelevance to our research objectives, despite appearing promising from their abstracts. Finally, we selected 26 papers from the secondary search (10 from the reference list and 16 from conference proceedings). This provided a total of 48 papers for this SLR for data extraction (see \autoref{fig:total_papers} and the full list of included papers in \autoref{appendix_A}).


\subsection{Quality Assessment}
In order to assess the quality of the selected papers, we employed the five-question assessment criteria proposed by Liu et al. \cite{liu2022systematic}. These questions assess the clarity of research aims, appropriateness of research design, clarity of findings and contributions, description of limitations and future work, and empirical nature of the study. Each question was evaluated on a scale of 0 to 1, with 0 indicating ``no'', 0.5 indicating ``partly'', and 1 indicating ``yes''. The overall quality score was calculated by summing the scores of the five questions, and the papers were classified as Good: if the score is between 3 and 4, Fair: if the score is between 2 and 3, Poor: if the score is between 0 and 2. Out of the 48 selected papers, 32 were deemed ``Good'' quality, 11 were ``Fair'' quality, and 5 were ``Poor'' quality, demonstrating the robustness of this review.


\subsection{Data Extraction}
\label{subsec:data_extraction}
Excel spreadsheet and NVivo software were used to extract demographic and content-related data from the 48 selected papers on D\&I in AI and AI for D\&I. The demographic data included the source of the paper, title, abstract, authors, affiliated countries of authors, year of publication, venue, and citation. Content-related data included the challenges faced to address D\&I in AI and AI for D\&I, and the proposed/used solutions (guidelines/ strategies/ approaches/ practices) to those challenges. The data was extracted through manual coding by the first author and cross-checked in weekly meetings with the other authors.


\subsection{Data Synthesis and Analysis}
\label{subsec:data_analysis}
The data synthesis and analysis for RQ1 and RQ2 is outlined in \autoref{fig:method}. To answer RQ1 and RQ2, the first author employed open coding to identify the challenges about D\&I in AI and AI for D\&I, as well as the proposed guidelines/ strategies/ approaches/ practices to address the challenges. All the authors checked the challenges and solutions to ensure inter-rater reliability and had several iteration of discussions to finalize them. The solutions were then mapped with the challenges for each paper to get a comprehensive understanding of what guidelines/ strategies/ approaches/ practices are taken for a specific challenge. Several iterations of discussions were conducted among all the authors to reach an agreement on the results for final mapping.
\section{Results}
\label{sec:results}

This section presents the results of the systematic literature review starting with the demographics of our selected 48 studies. We further present the extracted challenges of addressing diversity and inclusion in AI (D\&I in AI) and enhancing diversity and inclusion practices in the environment through AI (AI for D\&I), as well as the mentioned solutions to address the challenges. 


\subsection{Demographics}
Demographics covers a range of elements, including the publication year, citation count, whether the studies were empirical or theoretical in nature, the attributes of diversity addressed in each paper, as well as the affiliated countries of first authors.

    \begin{figure*}[!htbp]
        \centering
        \includegraphics[width=0.95\textwidth]{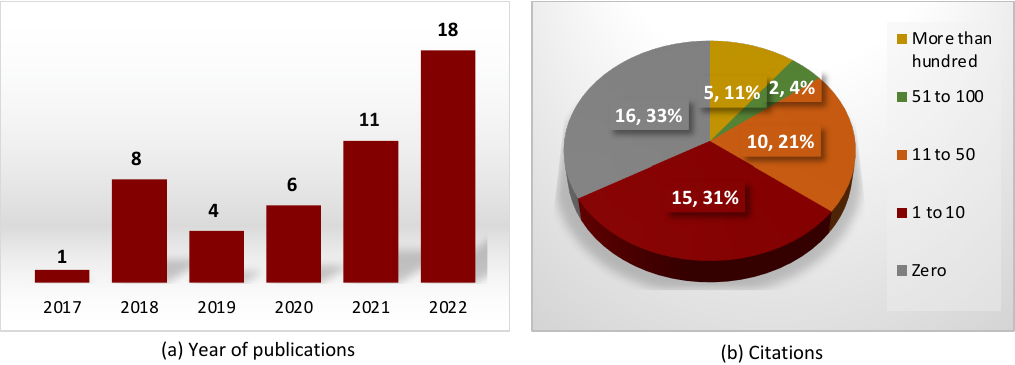}
        \caption{Year of publications and citations of the selected 48 papers}
        \label{fig:year_citation}
    \end{figure*}

The publication year and citations of the 48 selected papers are depicted in \autoref{fig:year_citation}. The data reveals that the majority of the papers (18) were published in 2022, followed by 11 papers published in 2021. Only one paper was published in 2017. This trend suggests that the field of D\&I in AI is relatively new, and further research in this area is needed. With regards to citations, \autoref{fig:year_citation} reveals that although we only covered last six years, five papers received more than 100 citations, while two papers received 51 to 100 citations. We also identified the number of empirical and non-empirical studies among the 48 selected studies. 30 of the selected studies are empirical and 18 are non-empirical.

The attributes of diversity analyzed in the selected studies, such as gender, age, and race, are depicted in \autoref{table:facets}. We differentiated the terms ``gender'' and ``sex'' in this table based on the terms used in the selected studies. According to Walker et al., ``Sex refers to
the anatomical or chromosomal categories of male and female. Gender refers to socially constructed roles that are related to sex distinctions'' \cite{walker1998brief}. The results suggest that the majority of the papers focus on gender (23 papers). There are also a good number of papers (15) on race, leaving room for further research on other attributes of diversity, such as age, sex, disability, neurodiversity, geographic location, skin tone, language, and ethnicity.

\begin{table*}
\centering
\caption{Attributes of diversity and their corresponding paper IDs}
\label{table:facets}
\resizebox{\textwidth}{!}{%
\renewcommand{\arraystretch}{1.05}
\begin{tabular}{|p{5.5cm}|p{3.2cm}|p{8.2cm}|}

\hline
\rowcolor[HTML]{D0CECE} 
\multicolumn{1}{|c|}{\cellcolor[HTML]{D0CECE}\textbf{Attributes}} & \textbf{Number of studies} & \multicolumn{1}{c|}{\cellcolor[HTML]{D0CECE}\textbf{Paper IDs}}                                                    \\ \hline
Gender                                                            & 23                           & S2, S3, S6, S10, S11,   S13, S15, S20, S21, S26, S27, S28, S29, S30, S31, S33, S36, S37, S39, S40,   S43, S45, S48 \\ \hline
Sex                                                               & 1                            & S10                                                                                                                \\ \hline
Age                                                               & 6                            & S6, S10, S15, S21,   S28, S33                                                                                      \\ \hline
Race                                                              & 15                           & S3, S6, S7, S10, S15,   S17, S20, S21, S23, S25, S26, S28, S33, S36, S41                                           \\ \hline
Ethnicity                                                         & 3                            & S3, S10, S26                                                                                                       \\ \hline
Disability                                                        & 4                            & S12, S14, S38, S42                                                                                                 \\ \hline
Neurodiversity                                                    & 1                            & S32                                                                                                                \\ \hline
Skin tone                                                         & 1                            & S13, S18, S25, S26                                                                                                 \\ \hline
Geographic location                                               & 2                            & S21, S35                                                                                                           \\ \hline
Family income and insurance status                                & 1                            & S25                                                                                                                \\ \hline
Language                                                          & 1                            & S33                                                                                                                \\ \hline

\end{tabular}
}
\end{table*}

    \begin{figure*}[!htbp]
        \centering
        \includegraphics[width=0.47\textwidth]{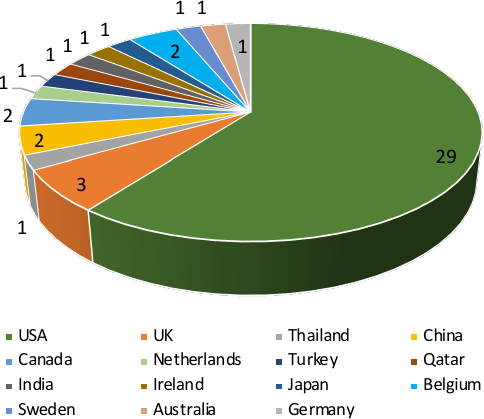}
        \caption{Affiliated countries of first authors of the selected papers}
        \label{fig:author_country}
    \end{figure*}

We also explored the ratio of affiliated countries of the first authors of the selected 48 studies which is presented in \autoref{fig:author_country}. The presence of United States of America (USA) is the maximum (29 out of 48), which reveals that the majority of the D\&I in AI or AI for D\&I related work has been conducted in USA. Three of the first authors are affiliated with United Kingdom (UK), two are affiliated with China, Canada, and Belgium each. Rest of the countries has only one occurrence each such as Thailand, Netherlands, Turkey, Qatar, India, Ireland, Japan, Sweden, Australia, and Germany. This findings reveal that diversity and inclusion in AI is the limited explored research area worldwide except USA. Therefore, this area should be focused more in future research.


\subsection{RQ1: Challenges and Solutions about Diversity and Inclusion in AI (D\&I in AI)}
\label{subsec:Ch_Sol_DI_AI}

\autoref{table:List_of_challenges_DI_AI} presents the list of challenges about D\&I in AI with their corresponding challenge IDs and paper IDs. We identified 55 unique challenges. Among the selected 48 papers, we identified challenges about D\&I in AI from 36 papers. We also identified 33 unique solutions to address some of those challenges as shown in \autoref{table:List_of_solutions_DI_AI}. Among the total of 48 papers, 23 papers discussed the solutions to the specific challenge mentioned. We also mapped the challenges with their corresponding solutions for each of the papers as presented in \autoref{appendix_C}. Some illustrative quotations on challenges and solutions for RQ1 as well some illustrative quotations on their mapping are presented below.

\begin{table*}
\centering
\caption{Results of RQ1: Challenges about diversity and inclusion in AI}
\label{table:List_of_challenges_DI_AI}
\resizebox{\textwidth}{!}{%
\renewcommand{\arraystretch}{1.05}
\begin{tabular}{|P{1cm}|p{13.5cm}|p{2.2cm}|}

\hline
\rowcolor[HTML]{E7E6E6} 
\multicolumn{1}{|c|}{\cellcolor[HTML]{E7E6E6}\textbf{Challenge ID}} & \multicolumn{1}{c|}{\cellcolor[HTML]{E7E6E6}\textbf{Challenges}}                                                                         & \multicolumn{1}{c|}{\cellcolor[HTML]{E7E6E6}\textbf{Paper ID}} \\ \hline
C1                                                                  & Concern about diversity without considering inclusion in AI                                                                              & S1                                                             \\ \hline
C2                                                                  & Lack of women role models in AI ecosystem                                                                                                & S2, S43                                                        \\ \hline
C3                                                                  & Lack of culture where women feel welcomed within the AI field                                                                            & S2                                                             \\ \hline
C4                                                                  & Lack of education on diversity, equity or disparities                                                                                    & S2                                                             \\ \hline
C5                                                                  & Socio-cultural norms and biases and stereotypes about women in AI                                                                        & S2                                                             \\ \hline
C6                                                                  & Lack of diversity in the composition of AI teams                                                                                         & S2, S10                                                        \\ \hline
C7                                                                  & Lack of Equity, Diversity and Inclusion (EDI) considerations in data set                                                                 & S5, S8, S10                                                    \\ \hline
C8                                                                  & Lack of diversity and inclusion issues in AI ethics documentation                                                                        & S5                                                             \\ \hline
C9                                                                  & Algorithm bias in human face processing technology                                                                                       & S7                                                             \\ \hline
C10                                                                 & Data handlers are influenced by their own backgrounds and prejudices                                                                     & S9                                                             \\ \hline
C11                                                                 & Certain communities' voices are disregarded and not uplifted in AI practice                                                              & S9, S36, S37, S44                                              \\ \hline
C12                                                                 & Bias in the training data sets                                                                                                           & S9, S17, S27                                                   \\ \hline
C13                                                                 & Implicit social bias in AI around ageism                                                                                                 & S10                                                            \\ \hline
C14                                                                 & Lack of diverse race, ethnicity, sex and gender inclusion and representation in the design, development, and implementation of AI system & S10                                                            \\ \hline
C15                                                                 & Lack of Equity, Diversity, and Inclusion (EDI) principles and indicators                                                                 & S10                                                            \\ \hline
C16                                                                 & Lack of a definitive and inclusive definition of Equity, Diversity, and Inclusion (EDI)                                                  & S10                                                            \\ \hline
C17                                                                 & Under-representation of minority groups in sampling during model training and testing                                                    & S10, S25, S42                                                  \\ \hline
C18                                                                 & Unconscious biases by AI, which is a manifestation of its creators' biases                                                               & S10, S15, S17, S29                                             \\ \hline
C19                                                                 & Unrealisticness of having universally inclusive dataset                                                                                  & S11                                                            \\ \hline
C20                                                                 & Lack of broader definitions of any given gender in gender classification system                                                                                         & S11                                                            \\ \hline
C21                                                                 & Absence of an explicit gender classifier in facial analysis (FA) services                                                                & S11                                                            \\ \hline
C22                                                                 & Biases in model design, training, or implementation                                                                                      & S11, S25                                                       \\ \hline
C23                                                                 & AI industry is dominated by men                                                                                                          & S15                                                            \\ \hline
C24                                                                 & Hiring AI developer and other technical roles is not gender-neutral                                                                      & S15                                                            \\ \hline
C25                                                                 & Inequalities in training data and lack of balanced inclusion data in machine algorithms                                                  & S15, S34                                                       \\ \hline
C26                                                                 & Insufficiency of "diversity" and "inclusion" related terminologies                                                                       & S16                                                            \\ \hline
C27                                                                 & Lack of AI researcher diversity                                                                                                          & S17                                                            \\ \hline
C28                                                                 & Inability to recognise the bias in data source due to lack of knowledge                                                                  & S17                                                            \\ \hline
C29                                                                 & Data invisibility, incomplete data, missing data                                                                                         & S17                                                            \\ \hline
C30                                                                 & Lack of equitable standards to address diversity in algorithm                                                                            & S17                                                            \\ \hline
C31                                                                 & Lack of comprehensive and accurate collection and generation of demographic data                                                         & S17                                                            \\ \hline
C32                                                                 & The underrepresentation of dark skin images in training data for diagnosing skin pathology                                               & S18                                                            \\ \hline
C33                                                                 & Lack of accuracy in unsupervised classification of diverse audiences                                                                     & S21                                                            \\ \hline
C34                                                                 & Traditional design processes, tools, and methods are difficult to effectively analyze and deal with complex diversity factors            & S22                                                            \\ \hline
C35                                                                 & Neglecting the level of diversity required by contextualized user-sensitive design                                                       & S22                                                            \\ \hline
C36                                                                 & Less attention on equity and justice principles in AI design and development                                                             & S23                                                            \\ \hline
C37                                                                 & Limited focus on diversity, equity, or disparities in AI-based academic literature                                                       & S24                                                            \\ \hline
C38                                                                 & Stakeholder roles and experiences are overlooked in equitable AI design, implementation, and use                                         & S24                                                            \\ \hline
C39                                                                 & Unstability of race and ethnic data labels due to the inconsistency of racial and ethnic categories across geographies                   & S26                                                            \\ \hline
C40                                                                 & Internal and subjective identities are not discussed when designing algorithms                                                           & S27                                                            \\ \hline
C41                                                                 & Subjective nature of gender identity                                                                                                     & S27                                                            \\ \hline
C42                                                                 & Algorithm learning discriminatory behavior from human behavior                                                                           & S29                                                            \\ \hline
C43                                                                 & Ad algorithms are not gender-neutral                                                                                                     & S29                                                            \\ \hline
C44                                                                 & Humans' unreliability of, or weakness in, spotting subconscious biases in AI systems                                                     & S30                                                            \\ \hline
C45                                                                 & Stereotypical gender concepts are embedded in the data                                                                                   & S31                                                            \\ \hline
C46                                                                 & Systematic imbalance of gender representation in the design of AI systems                                                                & S31, S36                                                       \\ \hline
C47                                                                 & Gender and racial discrimination in and by AI development team                                                                           & S33                                                            \\ \hline
C48                                                                 & Computer vision systems are not inclusive for all people from different demographics                                                     & S35                                                            \\ \hline
C49                                                                 & Overlooking disability considerations in ethical or legal levels of AI algorithms                                                        & S38                                                            \\ \hline
C50                                                                 & No comprehensive analysis of how gender is theorized in natural language processing                                                      & S39                                                            \\ \hline
C51                                                                 & Less examples of feminist and participatory methodologies to address power inequalities                                                  & S40                                                            \\ \hline
C52                                                                 & Racial categories are ill-defined in computer vision systems                                                                             & S41                                                            \\ \hline
C53                                                                 & Lack of data from disabled populations to train AI systems                                                                               & S42                                                            \\ \hline
C54                                                                 & Lack of trust in data accuracy in AI-enabled hiring software                                                                             & S46                                                            \\ \hline
C55                                                                 & Difficulties in measuring diversity in algorithm                                                                                         & S47                                                            \\ \hline

\end{tabular}
}
\end{table*}
\begin{table*}
\centering
\caption{Results of RQ1: Solutions to address the challenges of diversity and inclusion in AI}
\label{table:List_of_solutions_DI_AI}
\resizebox{\textwidth}{!}{%
\renewcommand{\arraystretch}{1.1}
\begin{tabular}{|p{1cm}|p{12.6cm}|p{1.4cm}|}

\hline
\rowcolor[HTML]{E7E6E6} 
\multicolumn{1}{|c|}{\cellcolor[HTML]{E7E6E6}\textbf{Solution ID}} & \multicolumn{1}{c|}{\cellcolor[HTML]{E7E6E6}\textbf{Solutions}}                                                                                                                                           & \multicolumn{1}{c|}{\cellcolor[HTML]{E7E6E6}\textbf{Paper ID}} \\ \hline
L1                                                                 & Implement awareness campaigns that tackle socio-cultural norms and biases and stereotypes                                                                                                                 & S2                                                             \\ \hline
L2                                                                 & Arrange unconscious bias training for teachers and counselors                                                                                                                                             & S2                                                             \\ \hline
L3                                                                 & Foster female-identifying role models in AI                                                                                                                                                               & S2, S31                                                        \\ \hline
L4                                                                 & Promote diversity and inclusion by developing methods and tools with diverse datasets that bring diversity and inclusion into engineering practice                                                        & S5                                                             \\ \hline
L5                                                                 & Use Lenovo face recognition engine (LeFace) to achieve better performance of racial fairness                                                                                                              & S7                                                             \\ \hline
L6                                                                 & Adopt data disaggregation by demographic groups                                                                                                                                                           & S8                                                             \\ \hline
L7                                                                 & Adopt more data examples in training for better learning outcomes                                                                                                                                         & S9                                                             \\ \hline
L8                                                                 & Adopt participatory design in AI system development, ensuring that all relevant stakeholders are represented/participated                                                                                 & S10, S27, S37                                                  \\ \hline
L9                                                                 & Develop policies to prevent discriminatory and nonconsensual gender representations in FA (Facial Analysis) systems                                                                                       & S11                                                            \\ \hline
L10                                                                & Consider purpose and definition of gender before gender classification in facial analysis or image labeling                                                                                               & S11                                                            \\ \hline
L11                                                                & Use diverse and inclusive data to train AI systems to be inclusive                                                                                                                                        & S15                                                            \\ \hline
L12                                                                & Remove discriminatory bias from job descriptions and resumes                                                                                                                                              & S15                                                            \\ \hline
L13                                                                & Configure AI to equitably screen candidates by disregarding age, gender, and race in profile assessment                                                                                                   & S15                                                            \\ \hline
L14                                                                & Use Word2Vec approach to visualize related terminologies                                                                                                                                                  & S16                                                            \\ \hline
L15                                                                & Establish accurate standards for the collection of detailed demographic data                                                                                                                              & S17                                                            \\ \hline
L16                                                                & Assess data used for models to avoid amplifying and perpetuating racial bias                                                                                                                              & S17                                                            \\ \hline
L17                                                                & Analyze publicly available skin image repositories to quantify the underrepresentation of darker skin tones                                                                                               & S18                                                            \\ \hline
L18                                                                & Enhance diversity-oriented design capacity to increase inclusiveness of diversity requirements                                                                                                            & S22                                                            \\ \hline
L19                                                                & Establish a user-centered machine learning system based on user and context features                                                                                                                      & S22                                                            \\ \hline
L20                                                                & Apply new tools, processes, and methods that algorithmically provide appropriate responses to specific needs                                                                                              & S22                                                            \\ \hline
L21                                                                & Designers require learning on diversity-oriented design                                                                                                                                                   & S22                                                            \\ \hline
L22                                                                & Integrate EDI (Equity, Diversity and Inclusion) and racial justice principles and practice in AI health                                                                                                   & S23                                                            \\ \hline
L23                                                                & Build a responsible culture in innovation and establish ethical building blocks for reliable delivery of equitable AI                                                                                     & S23                                                            \\ \hline
L24                                                                & Arrange training and education on the use of AI tools for equity promotion                                                                                                                                & S24                                                            \\ \hline
L25                                                                & Include diverse voices in training data and design                                                                                                                                                        & S25, S27                                                       \\ \hline
L26                                                                & Partner with ethicists and antiracism experts in developing, training, testing, and implementing models.                                                                                                  & S25                                                            \\ \hline
L27                                                                & Let users define their own gender identity before designing AI systems                                                                                                                                    & S27                                                            \\ \hline
L28                                                                & Use fairness indicators (e.g., harmful label association, geographical diversity and fairness, same-attribute assessment via similarity search) to probe main sources of biases in computer vision models & S35                                                            \\ \hline
L29                                                                & Use consistent, respectful, and accurate language for gender                                                                                                                                              & S39                                                            \\ \hline
L30                                                                & Use feminist research methodologies                                                                                                                                                                       & S39                                                            \\ \hline
L31                                                                & Adopt framework of data feminism to co-design datasets and machine learning models                                                                                                                        & S40                                                            \\ \hline
L32                                                                & Adopt fair computer vision datasets with different racial categories                                                                                                                                      & S41                                                            \\ \hline
L33                                                                & Adopt diversity by design, by operationalizing and implementing diversity-aware chatbot                                                                                                                   & S44                                                            \\ \hline

\end{tabular}
}
\end{table*}

\textbf{Illustrative Quotations on Challenges.}

\begin{itemize}
    \item [\faCommenting] Challenge C14: (Lack of diverse race, ethnicity, sex and gender inclusion and representation in the design, development, and implementation of AI system). \textit{``Lack of consideration for race, ethnicity, sex and gender in the design, development, and implementation of AI system in healthcare can lead to marginalization of underrepresented groups from benefitting from such technologies.''}- S10
\end{itemize}

\begin{itemize}
    \item [\faCommenting] Challenge C15: (The lack of Equity, Diversity, and Inclusion (EDI) principles and indicators). \textit{``The lack of EDI principles and indicators, for example, the presence of sex/gender, and racial/ethnicity bias in healthcare can be defined as differential medical and healthcare delivery and treatment of men, women, non-binary people and one race (dominant) over the others, the impact of which may be positive, negative, or neutral.''}- S10
\end{itemize}

\textbf{Illustrative Quotations on Solutions.}

\begin{itemize}
    \item [\faCommentsO] Solution L10: (Consider purpose and definition of gender before gender classification in facial analysis or image labeling). \textit{``Before embedding gender classification into a facial analysis service or incorporating gender into image labeling, it is important to consider what purpose gender is serving. Furthermore, it is important to consider how gender will be defined, and whether that perspective is unnecessarily exclusionary (e.g. binary).''}- S11 
\end{itemize}

\begin{itemize}
    \item [\faCommentsO] Solution L9: (Develop policies to prevent discriminatory and nonconsensual gender representations in FA (Facial Analysis) systems). \textit{``Establishing policies for how biometric data and face and body images are collected and used may be the most effective way of mitigating harm to trans people—and also people of marginalized races, ethnicities, and sexualities. Polices that prevent discriminatory and non-consensual gender representations could prevent gender misrepresentation from being incorporated into FA systems in both the data and infrastructure by regulating the use of gender as a category in algorithmic systems. For example, by banning the use of gender from FA-powered advertising and marketing.''}- S11 
\end{itemize}

\textbf{Illustrative Quotations on Mapping of Challenges and Solutions.}

\begin{itemize}  
    \item [\faCommenting] Challenge C52: (Racial categories are ill-defined in computer vision systems). \textit{``Racial categories are ill-defined, arbitrary and implicitly tied loosely to geographic origin. Second, given that racial categories are implicitly references to geographic origin, their extremely broad, continent-spanning construction would result in individuals with drastically different appearances and ethnic identities being grouped incongruously into the same racial category if the racial categories were interpreted literally. Thus racial categories must be understood both as references to geographic origin as well as physical characteristics.''}- S41
\end{itemize}

\begin{itemize}
    \item [\faCommentsO] Solution L32 to address C52: (Adopt fair computer vision datasets with different racial categories). \textit{``We empirically study the representation of race through racial categories in fair computer vision datasets, and analyze the crossdataset generalization of these racial categories, as well as their cross-dataset consistency, stereotyping, and self-consistency.''}- S41
\end{itemize}

\begin{figure*}[!htbp]
        \centering
        \includegraphics[width=1\textwidth]{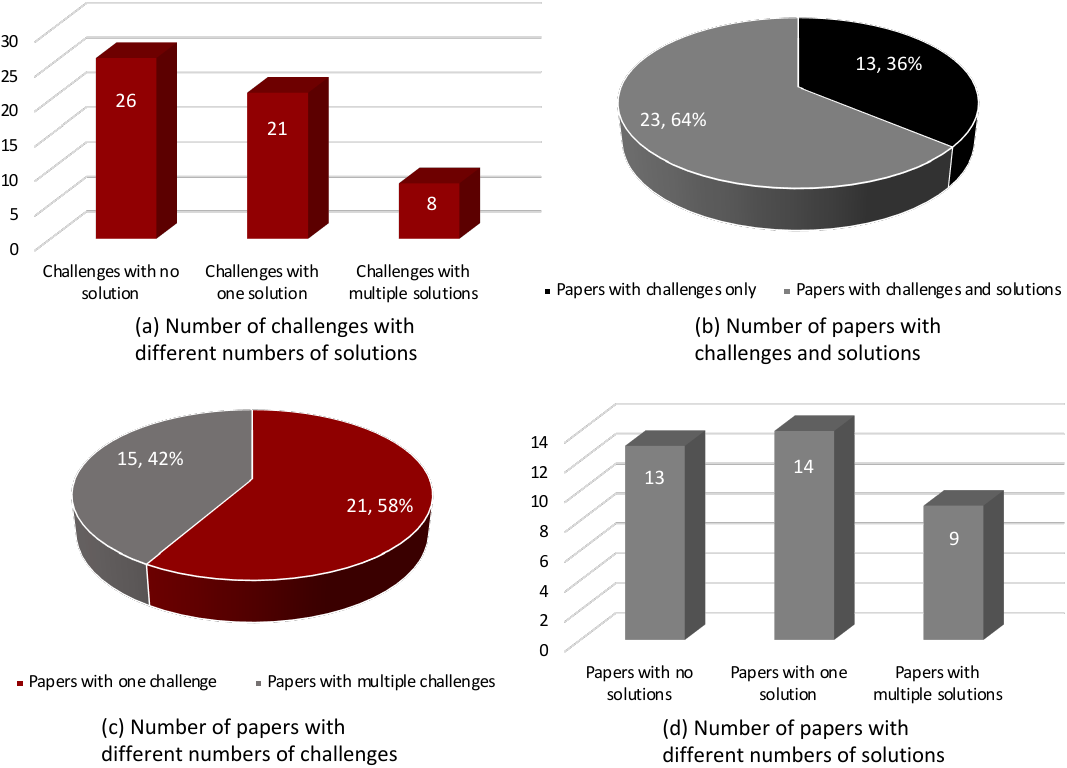}
        \caption{Analysis of the findings of RQ1}
        \label{fig:analysis_rq1}
    \end{figure*}

We have further analyzed to explore the findings from the mapping of challenges and solutions. According to \autoref{fig:analysis_rq1} (a), nearly half of the challenges (26) have no associated proposed solutions (e.g., C1, C3, C6, C23). 21 challenges have one solution each and the rest of the 8 challenges have more than one solutions. \autoref{fig:analysis_rq1} (b) shows the number of papers that have challenges with and without solutions. Among the 36 papers that identified the challenges about D\&I in AI (see \autoref{appendix_C}), 13 papers discussed about challenges with no solutions such as S1, S21, S42. The rest of the 23 papers discussed the challenges with possible solutions such as S2, S8, S23. \autoref{fig:analysis_rq1} (c) presents the number of papers with different numbers of challenges. More than half of the papers (21 papers) have only one challenge each (e.g., S1, S7, S47). The rest of the 15 papers mentioned more than one challenges such as S2, S10, S29. The last pie chart (see \autoref{fig:analysis_rq1} (d)) shows the ratio of papers with no solution, one solution and multiple solutions. Majority of the papers (14 papers) have one solution each (e.g., S7, S16, S40). However, there are a large number of papers (13 papers) which did not propose any solution at all such as S29, S36, S46. On the other hand, 9 papers proposed more than one solution for the challenges such as S2, S15, S22.

According to \autoref{appendix_C}, the paper S10 discussed the maximum number of challenges (8) about D\&I in AI such as ``Lack of Equity, Diversity, and Inclusion (EDI) principles and indicators'', ``Under-representation of minority groups in sampling during model training and testing''. The paper S17 also discussed a large number of challenges (7) such as ``Bias in the training data sets'' and ``Lack of comprehensive and accurate collection and generation of demographic data''. The paper S22 provided the maximum number of solutions (4) to address the challenges about D\&I in AI. For example, ``Enhance diversity-oriented design capacity to increase inclusiveness of diversity requirements'' and ``Establish a user-centered machine learning system based on user and context features''. Similarly, three papers (S2, S15, S27) provided three solutions each to address the challenges about D\&I in AI. Some of the papers provided multiple solutions for one challenge. For example, S22 provided three solutions for the challenges C35. Similarly, the papers S15, S23, S25 and S39 provided two solutions each to address one challenge (C24, C36, C22, C50 respectively).

We also identified some of the challenges which have been mentioned by more than one paper (see \autoref{table:List_of_challenges_DI_AI}). For example, C11 and C18 have been mentioned by four papers. Similarly, three papers discussed each of the challenges of C7, C12, and C17. Similar to the challenges, three solutions (L3, L8, L25) have been discussed by multiple papers (see \autoref{table:List_of_solutions_DI_AI}).


\subsection{RQ2: Challenges and Solutions about the Applications of AI for Diversity and Inclusion practices (AI for D\&I)}
\label{subsec:Ch_Sol_AI_DI_Practice}

20 out of 48 papers focused on the applications of AI for enhancing D\&I practices (AI for D\&I). \autoref{table:List_of_challenges_RQ2} presents the list of 24 unique challenges about AI for D\&I. The solutions to address the challenges with corresponding paper IDs are presented in \autoref{table:List_of_solutions_RQ2}, where we identified 23 solutions. The mapping of challenges with their corresponding solutions for each paper is shown in \autoref{appendix_D}. Some illustrative quotations on challenges and solutions for RQ2 as well some illustrative quotations on their mapping are presented below.

\clearpage
\begin{table*}
\centering
\caption{Results of RQ2: Challenges of the applications of AI for diversity and inclusion practices}
\label{table:List_of_challenges_RQ2}
\resizebox{\textwidth}{!}{%
\renewcommand{\arraystretch}{1.2}
\begin{tabular}{|p{1cm}|p{12.7cm}|p{1.3cm}|}

\hline
\rowcolor[HTML]{E7E6E6} 
\multicolumn{1}{|c|}{\cellcolor[HTML]{E7E6E6}\textbf{Challenge ID}} & \multicolumn{1}{c|}{\cellcolor[HTML]{E7E6E6}\textbf{Challenges}}                                                                                                                             & \multicolumn{1}{c|}{\cellcolor[HTML]{E7E6E6}\textbf{Paper ID}} \\ \hline
H1                                                                  & Lack of ability of facial recognition software to assess racial and ethnic diversity in qualitative medical studies                                                                          & S3                                                             \\ \hline
H2                                                                  & Bias by AI in workplace                                                                                                                                              & S4                                                             \\ \hline
 
H3                                                                  & AI based decisions exhibit discrimination based on sensitive attributes such as age, gender, and race                                                                                        & S6                                                             \\ \hline
H4                                                                  & Bias by AI in decisions related to hiring, compensation, and promotion                                                                                                                       & S8                                                             \\ \hline
H5                                                                  & Gender classifier is not used to mitigate gender bias                                                                                                                                        & S11                                                            \\ \hline

H6                                                                  & Underrepresented genders are not acknowledged by gender classification systems                                                                                                               & S11, S13                                                       \\ \hline
H7                                          & Inaccurate data label detection                                                                                                                                                              & S11                                    \\ \hline
H8                                          & Gender data labeling by gender classification systems offers limited labels to third-party developers                                                                                        & S11                                    \\ \hline
H9                                                                  & Lack of machine learning process to engage data by co-researchers with learning disabilities (LDs)                                                                   & S12                                                            \\ \hline
H10                                                                 & Bias by machine algorithms within a diverse pool of personnel                                                                                                                                & S15                                                            \\ \hline
H11                                                                 & Delayed or incorrect diagnoses of skin cancer for the people of color by early detection system                                                                                              & S18                                                            \\ \hline
H12                                                                 & Lack of use of machine learning technology in organizational diversity research                                                                                      & S19                                                            \\ \hline
 
H13                                                                 & AI replaces certain jobs that are predominantly held by underrepresented groups                                                                                                              & S20                                                            \\ \hline
H14                                                                 & Difficulties in understanding through AI how important Africans, women, and young people are in protecting, restoring, and promoting the sustainable use of terrestrial ecosystems in Africa & S21                                                            \\ \hline
H15                                                                 & Difficulties in identifying corresponding design patterns by machine learning technology after changing design requirements and problems                                                     & S22                                                            \\ \hline
H16                                                                 & Less accuracy of facial recognition technology to identify non-binary gender                                                                                                                      & S27                                                            \\ \hline
H17                                                                 & Difficulties in accuracy of diversity attributes detection in face detection tools                                                                                                           & S28                                                            \\ \hline
 
H18                                                                 & Lack of use of AI technology in delivering support to the children with autism                                                                                                               & S32                                                            \\ \hline
H19                                                                 & Difficulties in analysing disability requirements for AI in recruitment                                                                                                                      & S38                                                            \\ \hline
H20                                                                 & Bias in hiring with AI towards the people with disability                                                                                                                                    & S38                                                            \\ \hline
H21                                                                 & Disability is not widely studied in mitigation of bias in AI algorithms on ethical, legal or technical levels                                                                                & S38                                                            \\ \hline
H22                                                                 & Disability-based discrimination by AI technologies                                                                                                                                           & S42                                                            \\ \hline
H23                                                                 & Difficulties in estimating diversity in a given dataset                                                                                                                                      & S47                                                            \\ \hline
H24                                                                 & Lack of use of AI in understaing the diversity of people in any social media activism campaign                                                                       & S48                                                            \\ \hline

\end{tabular}
}
\end{table*}
\begin{table*}
\centering
\caption{Results of RQ2: Solutions of the challenges of AI applications for diversity and inclusion practices}
\label{table:List_of_solutions_RQ2}
\resizebox{\textwidth}{!}{%
\renewcommand{\arraystretch}{1.2}
\begin{tabular}{|p{1cm}|p{13cm}|p{1cm}|}

\hline
\rowcolor[HTML]{E7E6E6} 
\multicolumn{1}{|c|}{\cellcolor[HTML]{E7E6E6}\textbf{Solution ID}} & \multicolumn{1}{c|}{\cellcolor[HTML]{E7E6E6}\textbf{Solutions}}                                                                                                                                                                                   & \multicolumn{1}{c|}{\cellcolor[HTML]{E7E6E6}\textbf{Paper ID}} \\ \hline
N1                                                                 & Use Betaface (Betaface.com) facial analysis software to determine the diversity attributes                                                                                                                                & S3                                                             \\ \hline
N2                                                                 & Use AI to bring in the appropriate knowledge on bias reduction techniques and methods                                                                                                                                     & S4                                                             \\ \hline
N3                                                                 & Use AI to adopt fairness standards                                                                                                                                                                                        & S4                                                             \\ \hline
N4                                                                 & Use AI to make use of the relevant information and methods on data and algorithm development                                                                                                                              & S4                                                             \\ \hline
N5                                                                 & Use data analytics tools to improve decisions related to hiring, compensation, promotion, and retention, which can advance Diversity, Equity and Inclusion (DEI) practice                                                 & S8                                                             \\ \hline
N6                                                                 & Design Facial analysis services with the knowledge of the negative consequences of not recognizing genders correctly                                                                                                      & S11                                                            \\ \hline
N7                                                                 & Consider feature-based data labeling during design of Facial analysis services                                                                                                                                            & S11                                                            \\ \hline
N8                                                                 & Consider how to provide gender classification functionality to third-party developers with more scrutiny and oversight                                                                                                    & S11                                                            \\ \hline
N9                                                                 & Use qualitative data for labeling in gender classification                                                                                                                                                                                        & S11                                                            \\ \hline
N10                                                                & Adopt five-steps machine learning based structural topic modelling (STM) co-analysis process for creative, inclusive, and critical engagement of data by co-researchers                                                   & S12                                                            \\ \hline
 
N11                                                                & Assist in diverse candidate selection via social media by AI                                                                                                                                                                                      & S15                                                            \\ \hline
 
N12                                                                & Adopt AI-based skin cancer early detection system for all skin tones using clinical images                                                                                                                                                        & S18                                                            \\ \hline
N13                                                                & Adopt unsupervised machine-learning models to text mine and analyze how organizations communicate about Diversity, Equity and Inclusion (DEI) topics                                                                      & S19                                                            \\ \hline
N14                                                                & Adopt topic modeling to statistically identify word groups underlying all the diversity documents                                                                                                                         & S19                                                            \\ \hline
N15                                                                & Use a combination of self-reported location data, computer vision of social media photographs, natural language processing to estimate the demographics of individuals participating in the Global Landscapes Forum (GLF) & S21                                                            \\ \hline
N16                                                                & Adopt inclusive design tools, processes, and methods combined with machine learning technology to identify corresponding design patterns                                                                                  & S22                                                            \\ \hline
N17                                                                & Train automatic gender recognition (AGR) with a variety of gender identities early in the design process, by working with diverse teammembers and adopting participatory design approaches to identify non-binary gender  & S27                                                            \\ \hline
N18                                                                & Use face detection tools such as Face++, IBM Bluemix Visual Recognition, AWS Rekognition, and Microsoft Azure Face API to detect diversity attributes                                                                     & S28                                                            \\ \hline
 
N19                                                                & Use ECHOES that utilises an AI virtual character to facilitate autistic children's ability to engage in social interaction                                                                                                                        & S32                                                            \\ \hline
N20                                                                & Adopt analytical roadmap named ``recruitment AI'' to help mitigating the bias towards people with disability through ethical, legal and technical analysis                                                                  & S38                                                            \\ \hline
 
N21                                                                & Develop AI-powered accessibility tools to raise accessibility awareness in AI                                                                                                                                                                     & S42                                                            \\ \hline
N22                                                                & Adopt an approach called ``algorithmic greenlining'' to use diversity estimates instead of true diversity scores                                                                                                            & S47                                                            \\ \hline
N23                                                                & Use AI to analyze the diversity attributes from social media data                                                                                                                                                         & S48                                                            \\ \hline

\end{tabular}
}
\end{table*}


\textbf{Illustrative Quotation on Challenge.}

\begin{itemize}
    \item [\faCommenting] Challenge H6: (Underrepresented genders are not acknowledged by gender classification systems). \textit{``When classifying gender, designers of the systems we studied chose to use only two predefined demographic gender categories: male and female. As a result, these presentations are recorded, measured, classified, labeled, and databased for future iterations of binary gender classification.''}- S11 
\end{itemize}

\textbf{Illustrative Quotation on Solution.}

\begin{itemize}
    \item [\faCommentsO] Solution N1: (Use Betaface (Betaface.com) facial analysis software to determine the diversity attributes). \textit{``To determine the rates of diversity within departments, Betaface facial analysis software was used to analyze photos taken from the hospitals' websites. This software was able to determine the race, ethnicity, and gender of the care providers.''}- S3
\end{itemize}

\textbf{Illustrative Quotations on Mapping of Challenges and Solutions.}

\begin{itemize}  
    \item [\faCommenting] Challenge H16: (Less accuracy of facial recognition technology to identify non-binary gender). \textit{``This work which positions transgender faces as problematic to facial recognition accuracy, also raised ethical issues related to user privacy as the data for the database was scraped from transgender individuals’ videos without their consent or knowledge.''}- S27
\end{itemize}

\begin{itemize}
    \item [\faCommentsO] Solution N17 to address H16: (Train automatic gender recognition (AGR) with social and ethical implications). \textit{``While AGR technology is still in its infancy, the recent integration of facial recognition into already pervasive technologies suggest it could impact large numbers of people in the near future. As technologists continue to develop AGR applications, it is important to understand the social and ethical implications of widespread adoption.''}- S27
\end{itemize}


    \begin{figure*}[!htbp]
        \centering
        \includegraphics[width=1\textwidth]{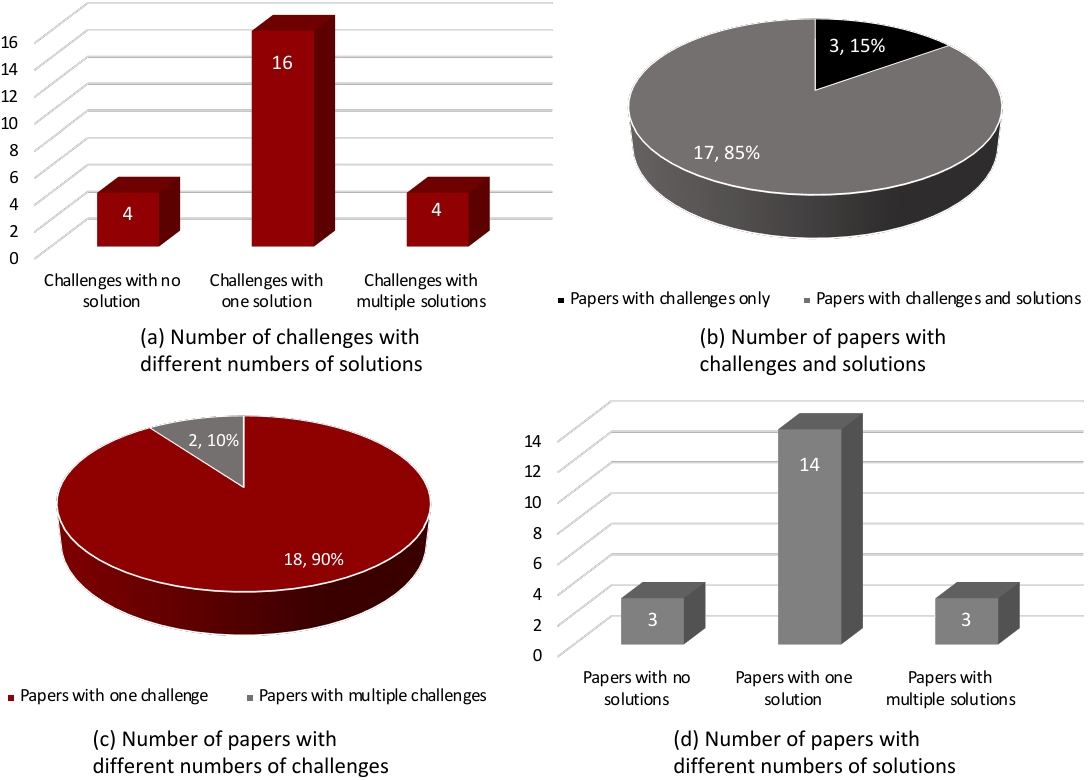}
        \caption{Analysis of the findings of RQ2}
        \label{fig:analysis_rq2}
    \end{figure*}

We have also explored additional findings from the mapping of challenges and solutions for RQ2, which is presented in \autoref{fig:analysis_rq2}. As shown in \autoref{fig:analysis_rq2}(a), the majority of the challenges (16) have one solution each such as H1, H10, H24. On the other hand, four challenges have no solution at all (H3, H5, H13, H19) and another four challenges have more than one solutions (H2, H7, H8, H12). Out of 20 papers on AI for D\&I, majority of the papers (17 papers) provided both challenges and solutions on the applications of AI for enhancing D\&I practices (see \autoref{fig:analysis_rq2} (b)). Three papers discussed challenges without any solutions. According to \autoref{fig:analysis_rq2} (c), two papers (S11, S38) discussed more than one challenge, whereas the rest of the 18 papers provided only one challenge each. S11 discussed the maximum number of challenges (4). As shown in \autoref{fig:analysis_rq2} (d), majority of the papers (14 out of 20) provided one solution each to address the challenges related to AI for D\&I. On the other hand, three papers did not propose any solution and another three papers provided multiple solutions for the challenges.


\subsection{Diversity Attributes}
\label{subsec:diversity_attributes}

\begin{figure*}[!htbp]
        \centering
        \includegraphics[width=1\textwidth]{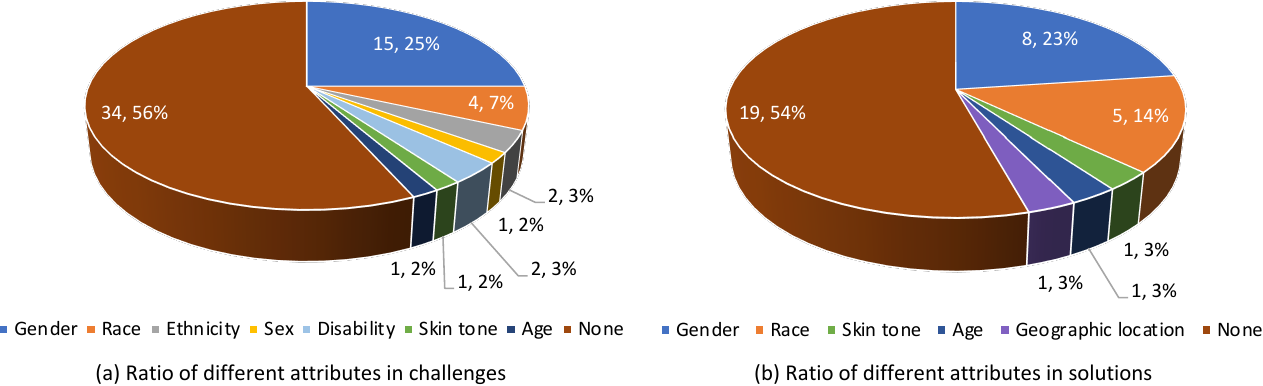}
        \caption{Ratio of diversity attributes in challenges and solutions for RQ1}
        \label{fig:visualization_attributes_DI_in_AI}
    \end{figure*}

\autoref{fig:visualization_attributes_DI_in_AI} illustrates the ratio of diversity attributes (e.g., age, gender, race, ethnicity) discussed in the challenges and solutions about D\&I in AI (RQ1). Majority of the challenges (56\%) and solutions (54\%) did not mention about any attributes at all. Gender has the maximum occurrences (25\% for challenges and 23\% for solutions). Race is the second highest attribute that was discussed in 7\% of the challenges and 14\% of the solutions.

    \begin{figure*}[!htbp]
        \centering
        \includegraphics[width=1\textwidth]{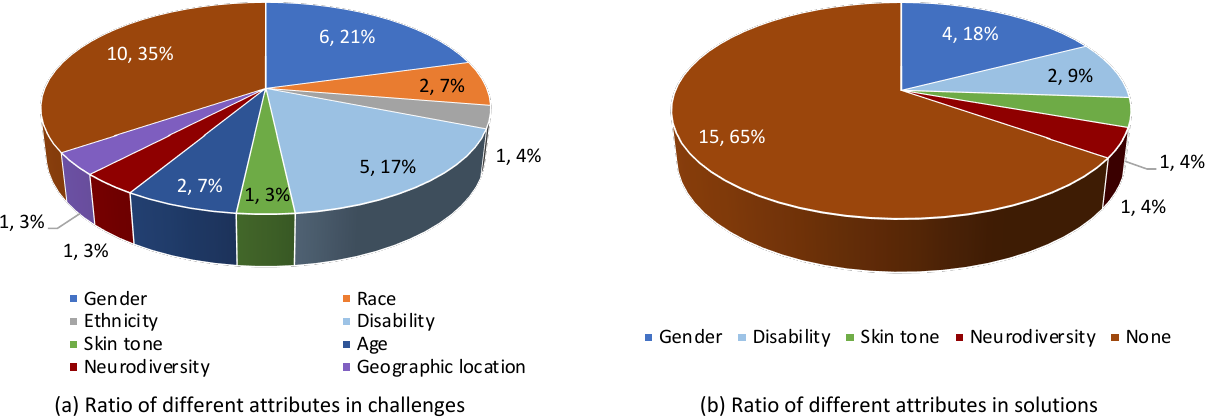}
        \caption{Ratio of diversity attributes in challenges and solutions for RQ2}
        \label{fig:visualization_attributes_AI_for_DI}
    \end{figure*}

\autoref{fig:visualization_attributes_AI_for_DI} illustrates the ratio of diversity attributes (e.g., age, gender, race, ethnicity) mentioned in the challenges and solutions about the applications of AI for D\&I practices (RQ2). According to \autoref{fig:visualization_attributes_AI_for_DI} (a), gender and disability are the two attributes that have the most occurrences (21\% for gender and 17\% for disability) in challenges. On the other hand, ethnicity, skin tone, neurodiversity, and geographic location have the least occurrences. \autoref{fig:visualization_attributes_AI_for_DI} (b) shows that majority of the solutions (65\%) do not indicate any attribute explicitly. However, gender has the maximum occurrences (18\%), whereas skin tone and neurodiversity have the least occurrences.

\section{Discussion and Implications}
\label{sec:discussion}

\subsection{Highlights of the Results}

\textbf{`Gender' as the Most Discussed Attribute of Diversity.}
Our analysis reveals that gender has been the top explored diversity attribute, in 23 out of the 48 papers (refer to \autoref{table:facets}). As for the other dimensions of diversity, 15 papers delved into race, 6 investigated age, 4 explored disability, 3 looked into ethnicity, 2 touched on geographic location, while sex, neurodiversity, skin tone, family income and insurance status, and language each took the spotlight in only a single paper. Moreover, when considering the challenges and solutions pertaining to D\&I in AI and AI for D\&I, gender has been the predominant topic of discussion (see \autoref{fig:visualization_attributes_DI_in_AI} and \autoref{fig:visualization_attributes_AI_for_DI}). For addressing D\&I in AI (RQ1), gender was the focus in 15 out of 55 challenges and 8 out of 33 solutions. Similarly, in the context of enhancing D\&I practices through AI (RQ2), 6 out of 24 challenges and 4 out of 23 solutions emphasized gender. The other diversity dimensions are largely overlooked. Recent studies \cite{roopaei2021women} have shed light on the challenges women face in AI, including bias, discrimination, a lack of self-confidence, inadequate resources and support, and limited exposure to AI in early education. Another study \cite{scheuerman2019computers} highlighted the difficulties faced by gender classifiers in recognizing non-binary genders. Despite these studies, there exists a dearth of research addressing other facets of diversity, like age, disability, race, ethnicity, and language. None of our included 48 studies worked on four of the attributes of diversity which was mentioned in the Article 26 of the International Covenant on Civil and Political Rights (ICCPR): religion, birth or other status, property, and national or social origin. Some recent federal laws of different countries such as Australian discrimination law \cite{australian2014quick} also discussed diversity attributes. None of our selected papers focused on many of the diversity attributes in Australian federal laws on discrimination such as religion, political opinion, and marital status. This underscores the necessity for more extensive investigations into these areas and the necessity to consider a broader spectrum of diversity, especially the notion of intersectionality, in both AI research and practice.

\begin{figure*}[!htbp]
        \centering
        \includegraphics[width=1\textwidth]{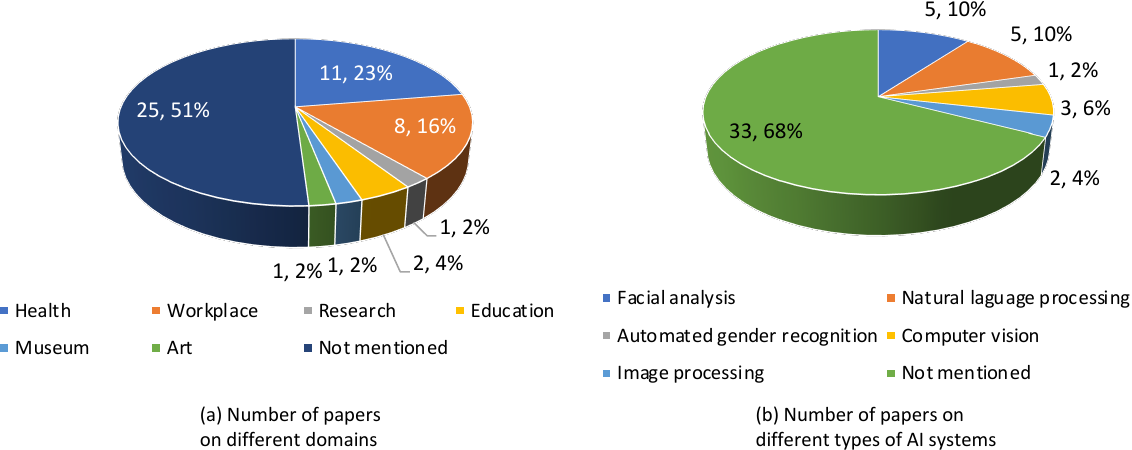}
        \caption{Ratio of different domains and types of AI systems}
        \label{fig:domain_type_of_AI}
    \end{figure*}

\textbf{`Health' as the Most Discussed Domain.}
Some of the included studies worked on D\&I in AI and AI for D\&I for some specific domains such as health, workplace, and education. \autoref{fig:domain_type_of_AI} (a) shows the ratio of different domains mentioned in the selected papers. More than half of the papers (51\%) do not focus on any specific domain, rather they discussed diversity and inclusion in AI in general. However, `health' is the most discussed domain, 23\% of the papers focused specifically on this domain. The second highest is `workplace' (16\%). Only a small number of papers mentioned about other domains such as `education', `research', `museum', and `art'. As many important domains such as law, banking, and transportation were not focused in any of the paper, more research is needed.

\textbf{`Facial Analysis' and `Natural Language Processing' as the Most Discussed Type of AI System.}
\autoref{fig:domain_type_of_AI} (b) illustrates the ratio of different types of AI systems, which were discussed in the selected studies of this SLR. Majority of the papers (68\%) did not mention any specific AI systems. Similar number of papers (10\%) discussed about facial analysis system and Natural language processing system. 6\% papers focused on computer vision system, 4\% on image processing, and 2\% on automated gender recognition system. Other types of AI systems must be studied with the lens of D\&I such as voice recognition and large language models.


\textbf{Lack of Solutions to Address D\&I in AI.} Number of solutions are less than the number of challenges about D\&I in AI (55 challenges, 33 solutions). \autoref{fig:analysis_rq1} (a) also shows that 26 out of 55 challenges have no solution to offer at all. Similarly, \autoref{fig:analysis_rq1} (b) shows that 36\% papers do not have any solution, whereas all of the papers discussed challenges. This finding leads to the necessity of further research in this area to recommend more solutions to address the challenges. Moreover, a large number (18 out of 48) of selected studies are non-empirical. This implies that proposed solutions are not implemented or validated in real settings. Therefore, there is a need for evidence-based research.




\textbf{Limited Research on AI for D\&I.} Our literature review shows that the majority of the selected papers (36 out of 48) discussed the challenges and some corresponding solutions to address D\&I in AI. On the other hand, a few papers (20 papers) discussed the challenges and solutions to enhance D\&I practices by AI (AI for D\&I). Similarly, the number of solutions to consider D\&I in AI is higher that the number of solutions to address AI for D\&I. The findings indicate that AI researchers are aware to address D\&I in AI, whereas AI for D\&I has taken limited attention. Although some recent studies worked on enhancing D\&I practices in workplace through AI \cite{chauhan2022role, pereira2023systematic} and enhancing D\&I practices in automated gender recognition systems \cite{scheuerman2019computers, hamidi2018gender}, further research needs to be conducted for more comprehensive understanding on AI for D\&I.


\textbf{Low Hanging Fruits.}
Our results revealed that various challenges could be tackled immediately with regard to diversity and inclusion in AI. For instance, including the perspectives of marginalized communities, such as individuals with disabilities and the elderly, in the development process, can support more representation in the training data \cite{huang2022social, nyariro2022integrating}. This can address various challenges, including the ``Under-representation of minority groups in sampling during model training and testing'' (S10, S25, S42), ``Certain communities' voices are disregarded and not uplifted in AI practice'' (S9, S36, S37, S44), ``Lack of comprehensive and accurate collection and generation of demographic data'' (S17), ``Overlooking disability considerations in ethical or legal levels of AI algorithms'' (S38). Additionally, promoting diversity in the recruitment of AI development teams and among researchers can help combat unconscious biases \cite{roopaei2021women, huang2022social, jora2022role, dankwa2022artificial}. Raising awareness and promoting education about diversity, equity, and disparities in AI can assist mitigating the knowledge gap about the people, places, and factors that make up the data \cite{roopaei2021women, dankwa2022artificial}.


\subsection{Five Pillars of Diversity and Inclusion in AI}

According to Zowghi and da Rimini, the definition of D\&I in AI consists of five pillars: \textit{Humans}, \textit{Data}, \textit{Process}, \textit{System}, and \textit{Governance} \cite{zowghi2023diversity}. We categorized our findings under the five pillars to explore the coverage of the challenges and solutions from this SLR within AI ecosystem and the pillars. We used these pillars for cross analysis and applied thematic coding on the findings for RQ1 and RQ2 to structure the challenges and solutions for D\&I in AI and AI for D\&I under the five pillars. It should be noted that the challenges and solutions for RQ1 and RQ2 are not necessarily mutually exclusive in relation to the five pillars. Therefore, many of them are listed under more than one pillar. This process was conducted independently by all of the authors and one external expert. An iterative series of discussions were conducted between all authors and the external annotator to ensure that the findings for answering RQ1 and RQ2 were accurately represented under their corresponding pillars. As all of the annotators have previous experience and expertise to this area and they analyzed the challenges and solutions from different disciplinary lens, we did not disregard any of their opinions. Therefore, we took the larger set which means we combined all the pillars categorized by all the annotators. The findings of our analysis are shown in \autoref{appendix_E}. Some examples of challenges and solutions for RQ1 and RQ2 under the five pillars are given below.

\begin{itemize}  
    \item [\faCommenting] \textbf{Humans}: (C11) Certain communities' voices are disregarded and not uplifted in AI practice. - S9, S36, S37, S44
    \item [\faCommenting] \textbf{Data}: (L6) Adopt data disaggregation by demographic groups. - S8
    \item [\faCommenting] \textbf{Process}: (H15) Difficulties in identifying corresponding design patterns by machine learning technology after changing design requirements and problems. - S22
    \item [\faCommenting] \textbf{System}: (N1) Use Betaface (Betaface.com) facial analysis software to determine the diversity attributes. - S3
    \item [\faCommenting] \textbf{Governance}: (C15) Lack of Equity, Diversity, and Inclusion (EDI) principles and indicators. - S10
\end{itemize}

\begin{figure*}[!htbp]
        \centering
        \includegraphics[width=1\textwidth]{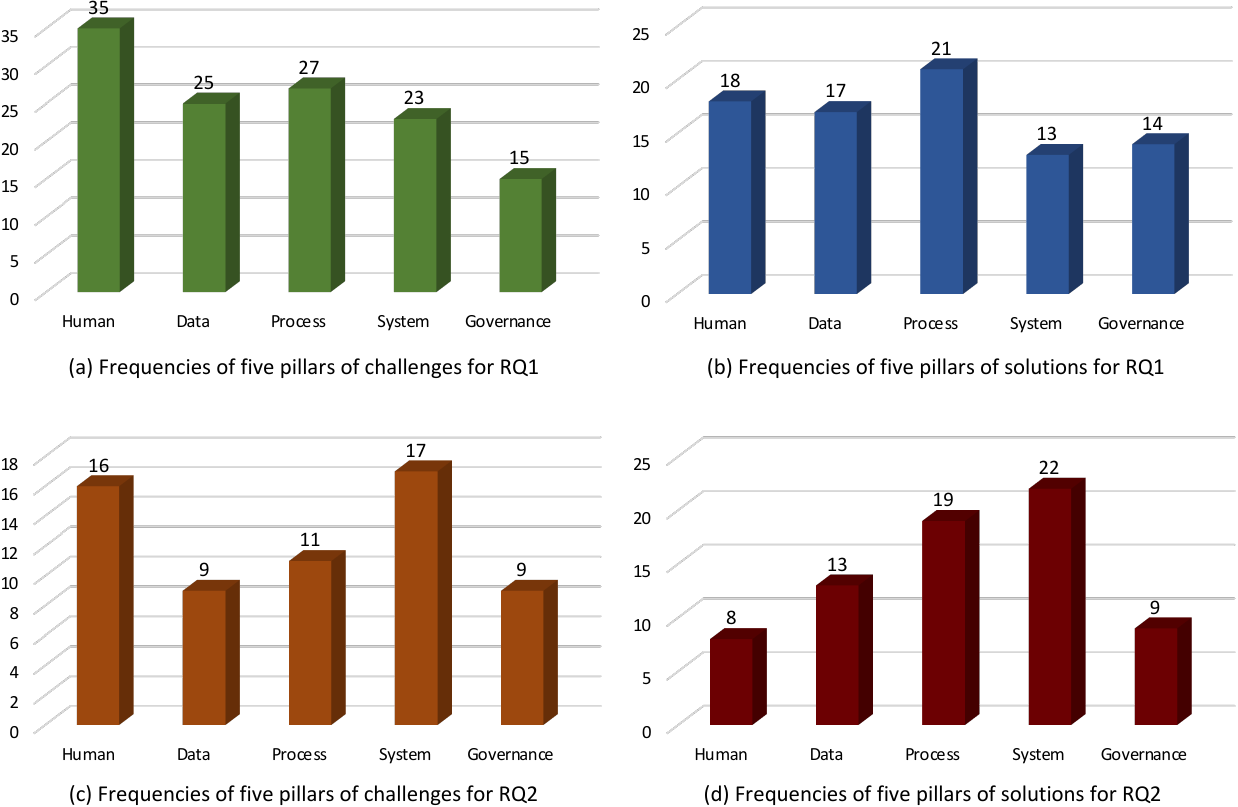}
        \caption{Frequencies of five pillars for the challenges and solutions for D\&I in AI (RQ1) and AI for D\&I (RQ2)}
        \label{fig:pillars}
    \end{figure*}

The frequencies of different pillars for the challenges and solutions for RQ1 (D\&I in AI) and RQ2 (AI for D\&I) are illustrated in \autoref{fig:pillars}. The findings revealed that \textit{Human}, not surprisingly, has the maximum occurrences for the challenges about D\&I in AI (RQ1) (see \autoref{fig:pillars} (a)). However, \textit{Process} is the highly addressed pillar in solutions to address the challenges about D\&I in AI (see \autoref{fig:pillars} (b)). \textit{System} is the most occurred pillar for both challenges and solutions for AI for D\&I (RQ2) (see \autoref{fig:pillars} (c) and (d)), whereas \textit{System} was mentioned less for RQ1.

On the other hand, we identified the least number of \textit{Governance} related challenges for both RQ1 and RQ2. \textit{Data} related challenges are also minimum for RQ2, whereas many challenges mentioned about \textit{Data} for RQ1. Surprisingly, \textit{Human} was mentioned the least for the solutions for RQ2.

As presented in \autoref{appendix_E}, the literature is limited in its coverage of diversity and inclusion in relation to governance of AI systems. Only a small number of studies mention the governance-related challenges and solutions associated with addressing D\&I in AI and AI for D\&I, such as ``lack of Equity, Diversity, and Inclusion (EDI) principles and indicators (C15)'', ``Integrate EDI (Equity, Diversity and Inclusion) and racial justice principles and practice in AI health (L22)'', ``Disability is not widely studied in mitigation of bias in AI algorithms on ethical, legal or technical levels (H21)'', ``Use AI to adopt fairness standards (N3)''. This is likely due to the fact that establishing D\&I principles and standards for AI systems often requires long-term planning, whereas addressing the challenges associated with humans, data, process, and systems can be addressed in less time. Therefore, it is crucial that policymakers be made aware of the importance of D\&I in AI in order to establish adequate plans for AI governance (such as standards, regulations, and policies) and principles to address these issues.


\subsection{Implications for Inclusive AI Systems Development}
In recent years, the importance of diversity and inclusion in AI and the corresponding have become increasingly acknowledged by researchers. Many challenges to address D\&I and AI for D\&I have been discussed in literature with various proposed solutions. One key solution is to raise awareness and provide training on cultural competency and algorithmic vigilance \cite{roopaei2021women, clark2021health}. This could help address socio-cultural norms, human biases, and stereotypes that may be embedded within AI systems \cite{roopaei2021women, dankwa2022artificial}. Another solution involves mitigating bias from job descriptions and resumes through training AI systems to disregard certain demographic information, such as age, gender, and race, while assessing profiles \cite{jora2022role}.

Inclusive design practices have also been suggested as a way to address D\&I in AI. This could involve adopting participatory design processes that involve diverse communities in the data collection and design process \cite{nyariro2022integrating}. Another approach involves combining inclusive design tools and methods with machine learning technology  to changes design requirements and identify corresponding design patterns \cite{li2020using}. Additionally, policy makers have a crucial role to play in addressing D\&I in AI. One suggestion is to establish more explicit policy documentation to ensure transparency on the policies \cite{huang2022social}.

Although we extracted and presented paper-wise solutions to address the challenges of D\&I in AI and AI for D\&I, some solutions from different papers could address a specific challenge. For example, the challenge, ``Underrepresented genders are not acknowledged by gender classification systems (H6)'' identified from the paper S11 and S13, could be addressed by the solutions from different papers such as ``Train automatic gender recognition (AGR) with a variety of gender identities early in the design process, by working with diverse teammembers and adopting participatory design approaches to identify non-binary gender (N17)'' (S27). This, along with other solutions, can help to ensure that AI systems are designed and developed in a manner that is inclusive and equitable for all.
\section{Threats to Validity}
\label{sec:ttv}

\textbf{Limitations.} 
Although we have rigorously adhered to the comprehensive search strategy dictated by the Evidence-Based SLR guidelines, ensuring a comprehensive selection of our samples, there's still a possibility that certain papers might not have been incorporated into our data collection. This may result from their inaccessibility or non-existence on electronic platforms, of which we might be unaware. \\
In the creation of our search strings, the key terms ``fairness'' and ``bias'' were deliberately omitted based on the insights from our pilot study and testing, with the objective of minimizing a large number of unrelated results. While we recognize this could have excluded certain relevant papers from our sample, we employed a meticulous secondary search strategy to counterbalance this limitation. This strategy, we believe, largely made up for the potential drawbacks of not using these terms initially. Nonetheless, we accept the possibility that some potentially relevant research might have been missed due to this strategic decision, though we stand firm in the overall effectiveness of our implemented research approach.

\textbf{Internal Validity.}
A potential threat could arise from the small number of selected papers and the restricted time span. As D\&I in AI and AI for D\&I are relatively new fields of research, we did not find many relevant papers prior to 2017. The majority of the papers were published recently (2022), and only 1 paper was published in 2017. However, in future studies, we will expand our time frame to check if there are more studies in this area. Another significant threat could arise from the bias in study selection and bias in data extraction. However, we mitigated this threat by adopting the investigator triangulation technique.

\textbf{Construct Validity.}
A potential construct threat could arise from the irrelevance of the selected papers with our research objectives. We selected many papers by reading the abstracts where there was a chance of getting information about D\&I in AI or AI for D\&I. However, many of them were removed after reading the full papers due to their irrelevance with our objectives. There is another potential threat to the subjective interpretation of the extracted data. Both of the threats were mitigated by adopting the investigator triangulation technique.

\textbf{External Validity.}
An external threat could arise from the generalizability of our findings. Although the results of this SLR may not be generalized for all types of AI technology, they can be considered representative within the specific domain of AI system development.
\section{Conclusions and Future Work}
\label{sec:conclusions}

We conducted a Systematic Literature Review with the goal to develop a comprehensive understanding of the challenges and corresponding solutions in addressing diversity and inclusion in artificial intelligence (D\&I in AI) and enhancing diversity and inclusion practices by artificial intelligence (AI for D\&I). After a rigorous process, we selected 48 academic papers published from 2017-2022, from which we extracted data and applied open coding on the data to explore information relevant to the challenges and solutions. Finally, we identified 55 unique challenges and 33 unique solutions in addressing D\&I in AI, and 24 unique challenges and 23 unique solutions in addressing AI for D\&I.

The analysis of the findings revealed that the integration of AI with diversity and inclusion is a less-explored area of research, as we found only a limited number of papers. Majority of these studies discussed the challenges of addressing D\&I in AI, but provided limited attention to the solutions to address those challenges. Moreover, a large number of solutions were proposed by some non-empirical studies without any implementation or validation in real life settings. Our study reveals that there is a lack of guidance for operationalizing the proposed solutions. We identified less challenges and solutions to address AI for D\&I from a limited number of papers compared to the number of challenges and solutions to address D\&I in AI. Hence, further research is required on AI for D\&I in particular and solutions of challenges for D\&I in AI.

Our results suggest that `Gender' is the most discussed attribute of diversity in AI, which leads to the necessity of further research on other attributes such as race, ethnicity, language, ageism, and religion. Similarly, `health' is the most discussed domain, and `facial analysis' and `natural language processing' are the most discussed AI systems in the analyzed literature on D\&I in AI and AI for D\&I, whereas other domains and types of AI systems are significantly ignored. We also identified that \textit{Governance} related issues are less discussed in the challenges and solutions to address D\&I in AI and AI for D\&I.

The results of our SLR have provided much-needed evidence for the advocacy of embedding and integrating D\&I practices and principles in the AI ecosystem. The gaps in the literature identified are the starting point for our holistic and comprehensive approach to tackling the D\&I-related issues in the overall AI ethics and Responsible AI body of knowledge. We have recognised the need for D\&I in AI guidelines and as a result, parallel to the conduct of this SLR, we have also performed a multi-vocal analysis of academic and grey literature to develop a comprehensive set of guidelines \cite{zowghi_hidden}. Our next step is to design and develop a risk-based framework for practitioners from the findings of this SLR that would incorporate a risk assessment checklist and context-specific recommendations for tackling the related issues at different stages of the AI development lifecycle. Our plan will include co-designing this framework by applying human-centred design and evidence-based approaches involving AI practitioners and relevant stakeholders.


\bibliography{References.bib}


\clearpage
\begin{appendices}





\clearpage
\section{List of 48 Included Studies}
\label{appendix_A}

\begin{table*}[h]
\centering
\resizebox{\textwidth}{!}{%
\renewcommand{\arraystretch}{0.8}
\begin{tabular}{p{1cm}p{15cm}}
\multirow{-0.2}{1cm}{S1}  & {\color[HTML]{222222} Mitchell,   Margaret, Dylan Baker, Nyalleng Moorosi, Emily Denton, Ben Hutchinson, Alex   Hanna, Timnit Gebru, and Jamie Morgenstern. ``Diversity and inclusion   metrics in subset selection.'' In Proceedings of   the AAAI/ACM Conference on AI, Ethics, and Society,   pp. 117-123. 2020.}                                                                                                                                 \\
\multirow{-0.2}{1cm}{S2}  & {\color[HTML]{222222} Roopaei,   Mehdi, Justine Horst, Emilee Klaas, Gwen Foster, Tammy J. Salmon-Stephens,   and Jodean Grunow. ``Women in ai: Barriers and solutions.'' In 2021 IEEE World AI IoT Congress (AIIoT), pp. 0497-0503. IEEE, 2021.}                                                                                                                                                                                                    \\
\multirow{-0.2}{1cm}{S3}  & {\color[HTML]{222222} Mathis,   Michelle S., Tosin E. Badewa, Ruth N. Obiarinze, Linda T. Wilkinson, and   Colin A. Martin. ``A novel use of artificial intelligence to examine   diversity and hospital performance.'' Journal of   Surgical Research 260 (2021): 377-382.}                                                                                                                                                                         \\
\multirow{-0.2}{1cm}{S4}  & {\color[HTML]{222222} Tongkachok,   Korakod, Shaifali Garg, Veena Prasad Vemuri, Vijesh Chaudhary, Poonam Vitthal   Koli, and K. Suresh Kumar. ``The Role of Artificial Intelligence on   Organisational support Programmes to Enhance work outcome and Employees   Behaviour.'' Materials Today: Proceedings 56 (2022): 2383-2387.}                                                                                                                 \\
\multirow{-0.2}{1cm}{S5}  & {\color[HTML]{222222} Chi, Nicole,   Emma Lurie, and Deirdre K. Mulligan. ``Reconfiguring diversity and   inclusion for AI ethics.'' In Proceedings of the   2021 AAAI/ACM Conference on AI, Ethics, and Society,   pp. 447-457. 2021.}                                                                                                                                                                                                              \\
\multirow{-0.2}{1cm}{S6}  & {\color[HTML]{222222} Srinivasan,   Ramya, and Kanji Uchino. ``The Role of Arts in Shaping AI Ethics.''   In AAAI Workshop on reframing diversity in AI:   Representation, inclusion, and power. CEUR Workshop Proceedings (CEUR-WS.   org). 2021.}                                                                                                                                                                                                  \\
\multirow{-0.2}{1cm}{S7}  & {\color[HTML]{222222} Shi, Sheng,   Shanshan Wei, Zhongchao Shi, Yangzhou Du, Wei Fan, Jianping Fan, Yolanda   Conyers, and Feiyu Xu. ``Algorithm Bias Detection and Mitigation in   Lenovo Face Recognition Engine.'' In Natural   Language Processing and Chinese Computing: 9th CCF International Conference,   NLPCC 2020, Zhengzhou, China, October 14–18, 2020, Proceedings, Part II 9, pp. 442-453. Springer International Publishing, 2020.} \\

\multirow{-0.2}{1cm}{S8}  & {\color[HTML]{222222} Chauhan,   Preeti S., and Nir Kshetri. ``The Role of Data and Artificial   Intelligence in Driving Diversity, Equity, and Inclusion.'' Computer 55, no. 4 (2022):   88-93.}                                                                                                                                                                                                             \\
\multirow{-0.2}{1cm}{S9} & {\color[HTML]{222222} Huang,   Han-Yin, and Cynthia CS Liem. ``Social Inclusion in Curated Contexts:   Insights from Museum Practices.'' In 2022 ACM   Conference on Fairness, Accountability, and Transparency, pp. 300-309. 2022.}                                                                                                                                                                                                                 \\
\multirow{-0.2}{1cm}{S10} & {\color[HTML]{222222} Nyariro,   Milka, Elham Emami, and Samira Abbasgholizadeh Rahimi. ``Integrating   Equity, Diversity, and Inclusion throughout the lifecycle of Artificial   Intelligence in health.'' In 13th Augmented   Human International Conference, pp. 1-4. 2022.}                                                                                                                                                                      \\
\multirow{-0.2}{1cm}{S11} & {\color[HTML]{222222} Scheuerman,   Morgan Klaus, Jacob M. Paul, and Jed R. Brubaker. ``How computers see   gender: An evaluation of gender classification in commercial facial analysis   services.'' Proceedings of the ACM on   Human-Computer Interaction 3, no. CSCW   (2019): 1-33.}                                                                                                                                                           \\
\multirow{-0.2}{1cm}{S12} & {\color[HTML]{222222} Chapko,   Dorota, Pedro Andrés Andrés Pérez Rothstein, Lizzie Emeh, Pino Frumiento,   Donald Kennedy, David McNicholas, Ifeoma Orjiekwe et al. ``Supporting   Remote Survey Data Analysis by Co-researchers with Learning Disabilities   through Inclusive and Creative Practices and Data Science Approaches.'' In Designing Interactive Systems Conference 2021, pp. 1668-1681. 2021.}                                     \\
\multirow{-0.2}{1cm}{S13} & {\color[HTML]{222222} Celis, L.   Elisa, and Vijay Keswani. ``Implicit diversity in image   summarization.'' Proceedings of the ACM on   Human-Computer Interaction 4, no. CSCW2   (2020): 1-28.}  
\\
\multirow{-0.2}{1cm}{S14} & {\color[HTML]{222222} Kurnaz, Sefer,   and Maalim AH Aljabery. ``Predict the type of hearing aid of audiology   patients using data mining techniques.'' In Proceedings   of the Fourth International Conference on Engineering \& MIS 2018, pp. 1-6. 2018.}                                                                                                                                                     
\end{tabular}
}
\end{table*}
\begin{table*}[h]
\centering
\resizebox{\textwidth}{!}{%
\renewcommand{\arraystretch}{0.8}
\begin{tabular}{p{1cm}p{15cm}}

\multirow{-0.2}{1cm}{S15} & {\color[HTML]{222222} Jora, Rachna   Bansal, Kavneet Kaur Sodhi, Prabhat Mittal, and Parul Saxena. ``Role of   Artificial Intelligence (AI) In meeting Diversity, Equality and Inclusion   (DEI) Goals.'' In 2022 8th International   Conference on Advanced Computing and Communication Systems (ICACCS), vol. 1, pp. 1687-1690. IEEE, 2022.} 
\\

\multirow{-0.2}{1cm}{S16} & {\color[HTML]{222222} Bhaduri,   Sreyoshi, and Tamoghna Roy. ``A word-space visualization approach to   study college of engineering mission statements.'' In 2017 IEEE Frontiers in Education Conference (FIE), pp. 1-5. IEEE, 2017.} 
\\

\multirow{-0.2}{1cm}{S17} & {\color[HTML]{222222} Dankwa-Mullan,   Irene, and Dilhan Weeraratne. ``Artificial intelligence and machine   learning technologies in cancer care: Addressing disparities, bias, and data   diversity.'' Cancer Discovery 12, no. 6 (2022): 1423-1427.} \\

\multirow{-0.2}{1cm}{S18} & {\color[HTML]{222222} Rezk, Eman,   Mohamed Eltorki, and Wael El-Dakhakhni. ``Leveraging Artificial   Intelligence to Improve the Diversity of Dermatological Skin Color Pathology:   Protocol for an Algorithm Development and Validation Study.'' JMIR Research Protocols 11, no.   3 (2022): e34896.}                                                                                                                                             \\
\multirow{-0.2}{1cm}{S19} & {\color[HTML]{222222} Wang, Wei, Julie V.   Dinh, Kisha S. Jones, Siddharth Upadhyay, and Jun Yang. ``Corporate   diversity statements and employees' online DEI ratings: An unsupervised   machine-learning text-mining analysis.'' Journal of Business and   Psychology (2022): 1-17.} \\

\multirow{-0.2}{1cm}{S20} & {\color[HTML]{222222} Ozkazanc-Pan,   Banu. ``Diversity and future of work: inequality abound or opportunities   for all?.'' Management Decision 59, no. 11 (2021): 2645-2659.}                                                                                                                                                                                                                                                                                                         \\
\multirow{-0.2}{1cm}{S21} & {\color[HTML]{222222} Brandt, John,   Kathleen Buckingham, Cody Buntain, Will Anderson, Sabin Ray, John-Rob Pool,   and Natasha Ferrari. ``Identifying social media user demographics and   topic diversity with computational social science: a case study of a major   international policy forum.'' Journal of   Computational Social Science 3 (2020):   167-188.}                                                                                                                  \\
\multirow{-0.2}{1cm}{S22} & {\color[HTML]{222222} Li, Fang, Hua   Dong, and Long Liu. ``Using AI to Enable Design for Diversity: A   Perspective.'' In Advances in Industrial Design:   Proceedings of the AHFE 2020 Virtual Conferences on Design for Inclusion,   Affective and Pleasurable Design, Interdisciplinary Practice in Industrial   Design, Kansei Engineering, and Human Factors for Apparel and Textile Engineering,   July 16–20, 2020, USA, pp. 77-84. Springer   International Publishing, 2020.} \\

\multirow{-0.2}{1cm}{S23} & {\color[HTML]{222222} Dankwa-Mullan,   Irene, Elisabeth Lee Scheufele, Michael E. Matheny, Yuri Quintana, Wendy W.   Chapman, Gretchen Jackson, and Brett R. South. ``A proposed framework on   integrating health equity and racial justice into the artificial intelligence   development lifecycle.'' Journal of Health Care   for the Poor and Underserved 32, no. 2   (2021): 300-317.}
\\
\multirow{-0.2}{1cm}{S24} & {\color[HTML]{222222} Clark, Cheryl   R., Consuelo Hopkins Wilkins, Jorge A. Rodriguez, Anita M. Preininger, Joyce   Harris, Spencer DesAutels, Hema Karunakaram, Kyu Rhee, David W. Bates, and   Irene Dankwa-Mullan. ``Health care equity in the use of advanced   analytics and artificial intelligence technologies in primary   care.'' Journal of General Internal Medicine 36 (2021): 3188-3193.}\\

\multirow{-0.2}{1cm}{S25} & {\color[HTML]{222222} Sveen,   William, Maya Dewan, and Judith W. Dexheimer. ``The Risk of Coding Racism   into Pediatric Sepsis Care: The Necessity of Antiracism in Machine   Learning.'' The Journal of Pediatrics 247 (2022): 129-132.}                                                                                                                                                                                                                                             \\
\multirow{-0.2}{1cm}{S26} & {\color[HTML]{222222} Buolamwini,   Joy, and Timnit Gebru. ``Gender shades: Intersectional accuracy   disparities in commercial gender classification.'' In Conference on fairness, accountability and transparency, pp. 77-91. PMLR, 2018.}                                                                                                                                                                                                                                            \\
\multirow{-0.2}{1cm}{S27} & {\color[HTML]{222222} Hamidi, Foad,   Morgan Klaus Scheuerman, and Stacy M. Branham. ``Gender recognition or   gender reductionism? The social implications of embedded gender recognition   systems.'' In Proceedings of the 2018 chi   conference on human factors in computing systems,   pp. 1-13. 2018.}                                                                                                                                                                           \\
\multirow{-0.2}{1cm}{S28} & {\color[HTML]{222222} Jung,   Soon-Gyo, Jisun An, Haewoon Kwak, Joni Salminen, and Bernard Jim Jansen.  ``Assessing the accuracy of four popular face recognition tools for   inferring gender, age, and race.'' In Twelfth   international AAAI conference on web and social media. 2018.}                                                                                                                                                                                            \\
\multirow{-0.2}{1cm}{S29} & {\color[HTML]{222222} Lambrecht,   Anja, and Catherine E. Tucker. ``Algorithmic bias? An empirical study   into apparent gender-based discrimination in the display of STEM career   ads.'' An Empirical Study into Apparent   Gender-Based Discrimination in the Display of STEM Career Ads (March 9, 2018) (2018).}                                                                                                                                                                   \\
\multirow{-0.2}{1cm}{S30} & {\color[HTML]{222222} Cohen, Tammy.   ``How to leverage artificial intelligence to meet your diversity   goals.'' Strategic HR Review (2019).}   \\

\multirow{-0.2}{1cm}{S31} & {\color[HTML]{222222} Leavy, Susan.   ``Gender bias in artificial intelligence: The need for diversity and   gender theory in machine learning.'' In Proceedings   of the 1st international workshop on gender equality in software engineering, pp. 14-16. 2018.}                                     

\end{tabular}
}
\end{table*}
\begin{table*}[h]
\centering
\resizebox{\textwidth}{!}{%
\renewcommand{\arraystretch}{0.8}
\begin{tabular}{p{1cm}p{15cm}}

\multirow{-0.2}{1cm}{S32} & {\color[HTML]{222222} Porayska-Pomsta,   Kaśka, Alyssa M. Alcorn, Katerina Avramides, Sandra Beale, Sara Bernardini,   Mary Ellen Foster, Christopher Frauenberger et al. ``Blending human and   artificial intelligence to support autistic children’s social communication   skills.'' ACM Transactions on Computer-Human   Interaction (TOCHI) 25, no. 6 (2018): 1-35.}                                                                                                              \\
\multirow{-0.2}{1cm}{S33} & {\color[HTML]{222222} Kong, Youjin.   ``Are ``Intersectionally Fair'' AI Algorithms Really Fair to Women of   Color? A Philosophical Analysis.'' In 2022 ACM   Conference on Fairness, Accountability, and Transparency, pp. 485-494. 2022.}                                                                                                                                                                                                                                              \\
\multirow{-0.2}{1cm}{S34} & {\color[HTML]{222222} Salem, Jad,   Deven Desai, and Swati Gupta. ``Don't let Ricci v. DeStefano Hold You   Back: A Bias-Aware Legal Solution to the Hiring Paradox.'' In 2022 ACM Conference on Fairness, Accountability, and   Transparency, pp. 651-666. 2022.}                                                                                                                                                                                                                      \\
\multirow{-0.2}{1cm}{S35} & {\color[HTML]{222222} Goyal, Priya,   Adriana Romero Soriano, Caner Hazirbas, Levent Sagun, and Nicolas Usunier.   ``Fairness indicators for systematic assessments of visual feature   extractors.'' In 2022 ACM Conference on   Fairness, Accountability, and Transparency, pp.   70-88. 2022.}                                                                                                                                                                                       \\
\multirow{-0.2}{1cm}{S36} & {\color[HTML]{222222} Hirota,   Yusuke, Yuta Nakashima, and Noa Garcia. ``Gender and racial bias in   visual question answering datasets.'' In 2022   ACM Conference on Fairness, Accountability, and Transparency, pp. 1280-1292. 2022.}                                                                                                                                                                                                                                               \\
\multirow{-0.2}{1cm}{S37} & {\color[HTML]{222222} Vlasceanu,   Madalina, Miroslav Dudik, and Ida Momennejad. ``Interdisciplinarity,   Gender Diversity, and Network Structure Predict the Centrality of AI   Organizations.'' In 2022 ACM Conference on   Fairness, Accountability, and Transparency, pp.   1-10. 2022.}                                                                                                                                                                                            \\
\multirow{-0.2}{1cm}{S38} & {\color[HTML]{222222} Buyl, Maarten,   Christina Cociancig, Cristina Frattone, and Nele Roekens. ``Tackling   Algorithmic Disability Discrimination in the Hiring Process: An Ethical,   Legal and Technical Analysis.'' In 2022 ACM   Conference on Fairness, Accountability, and Transparency, pp. 1071-1082. 2022.}                                                                                                                                                                  \\
\multirow{-0.2}{1cm}{S39} & {\color[HTML]{222222} Devinney,   Hannah, Jenny Björklund, and Henrik Björklund. ``Theories of ``gender'' in   NLP bias research.'' In 2022 ACM Conference on   Fairness, Accountability, and Transparency, pp.   2083-2102. 2022.}   \\

\multirow{-0.2}{1cm}{S40} & {\color[HTML]{222222} Suresh,   Harini, Rajiv Movva, Amelia Lee Dogan, Rahul Bhargava, Isadora Cruxen,   Ángeles Martinez Cuba, Guilia Taurino, Wonyoung So, and Catherine D'Ignazio.   ``Towards Intersectional Feminist and Participatory ML: A Case Study in   Supporting Feminicide Counterdata Collection.'' In 2022 ACM Conference on Fairness, Accountability, and   Transparency, pp. 667-678. 2022.} \\

\multirow{-0.2}{1cm}{S41} & {\color[HTML]{222222} Khan, Zaid,   and Yun Fu. ``One label, one billion faces: Usage and consistency of   racial categories in computer vision.'' In Proceedings   of the 2021 acm conference on fairness, accountability, and transparency, pp. 587-597. 2021.}                                                                                                                                             \\
\multirow{-0.2}{1cm}{S42} & {\color[HTML]{222222} Park, Joon   Sung, Danielle Bragg, Ece Kamar, and Meredith Ringel Morris. ``Designing   an online infrastructure for collecting AI data from people with   disabilities.'' In Proceedings of the 2021 ACM   Conference on Fairness, Accountability, and Transparency, pp. 52-63. 2021.}                                                                                                 \\
\multirow{-0.2}{1cm}{S43} & {\color[HTML]{222222} Cheong, Marc,   Kobi Leins, and Simon Coghlan. ``Computer science communities: Who is   speaking, and who is listening to the women? Using an ethics of care to   promote diverse voices.'' In Proceedings of the   2021 ACM Conference on Fairness, Accountability, and Transparency, pp. 106-115. 2021.}                                                                              \\
\multirow{-0.2}{1cm}{S44} & {\color[HTML]{222222} Helm, Paula,   Loizos Michael, and Laura Schelenz. ``Diversity by Design? Balancing the   Inclusion and Protection of Users in an Online Social Platform.''   In Proceedings of the 2022 AAAI/ACM Conference on   AI, Ethics, and Society, pp. 324-334. 2022.}                                                                                                                          \\
\multirow{-0.2}{1cm}{S45} & {\color[HTML]{222222} Siapka,   Anastasia. ``Towards a Feminist Metaethics of AI.'' In Proceedings of the 2022 AAAI/ACM Conference on AI, Ethics, and   Society, pp. 665-674. 2022.}                                                                                                                                                                                                                          \\
\multirow{-0.2}{1cm}{S46} & {\color[HTML]{222222} Li, Lan, Tina   Lassiter, Joohee Oh, and Min Kyung Lee. ``Algorithmic hiring in practice:   Recruiter and HR Professional's perspectives on AI use in hiring.''   In Proceedings of the 2021 AAAI/ACM Conference on   AI, Ethics, and Society, pp. 166-176. 2021.}                                                                                                                      \\
\multirow{-0.2}{1cm}{S47} & {\color[HTML]{222222} Borgs, Christian,   Jennifer Chayes, Nika Haghtalab, Adam Tauman Kalai, and Ellen Vitercik.   ``Algorithmic greenlining: An approach to increase diversity.'' In   Proceedings of the 2019 AAAI/ACM Conference on AI, Ethics, and Society, pp.   69-76. 2019.}                                                                                                                          \\
\multirow{-0.2}{1cm}{S48} & {\color[HTML]{222222} Karbasian,   Habib, Hemant Purohit, Rajat Handa, Aqdas Malik, and Aditya Johri.   ``Real-time inference of user types to assist with more inclusive and   diverse social media activism campaigns.'' In Proceedings   of the 2018 AAAI/ACM Conference on AI, Ethics, and Society, pp. 171-177. 2018.}

\end{tabular}
}
\end{table*}

\clearpage

\section{Search string in digital libraries and their corresponding results for pilot and main study}
\label{appendix_B}

\textbf{Pilot study}
\begin{table*}[h]
\centering
\label{table:pilot_search_string}
\resizebox{\textwidth}{!}{%
\renewcommand{\arraystretch}{1.3}
\begin{tabular}{|p{1.8cm}|P{5.5cm}|P{2.3cm}|P{1.4cm}|p{1.5cm}|p{1.4cm}|p{1.8cm}|}

\hline
\textbf{Digital Library}     & \textbf{Search String}                                                                                                                                                                                                                        & \textbf{Search Within}    & \textbf{Time Frame} & \textbf{Search Time}   & \textbf{No. of Papers Returned} & \textbf{Selected by Reading Abstract} \\ \hline
ACM Digital Library & {[}{[}Abstract: ``artificial intelligence''{]} OR {[}Abstract: ``machine learning''{]}{]} AND {[}Abstract: ``diversity and inclusion''{]} & Research article (Abstract)          & 2017 - 2022         & 15/05/2022 (2:47 am)    & 2  & 2                   \\ \hline
IEEE Xplore         & (``Abstract'':``artificial intelligence'' OR ``Abstract'':``machine learning'') AND (``Abstract'':``diversity and inclusion'')                                                             & Conference, Journal (Abstract)       & 2017 - 2022         & 15/05/2022 (2:47 am)    & 2  & 2                     \\ \hline
Science Direct      & (``artificial intelligence'' OR ``machine learning'') AND (``diversity and inclusion'')                                                                                                                         & Research articles (Title, Abstract, Keywords)         & 2017 - 2022         & 15/05/2022 (2:58 am)  & 8 & 2                     \\ \hline
Scopus              & ( TITLE-ABS-KEY (``artificial intelligence''  OR  ``machine learning'')  AND  TITLE-ABS-KEY (``diversity and inclusion'') )                                                                                   & Conference paper, Article (Title, Abstract, Keywords) & 2017 - 2022         & 15/05/2022 (1:02 pm)  & 13 & 6                     \\ \hline

Google Scholar              & allintitle: artificial AND intelligence AND diversity AND inclusion                                                                                   & All (Title, Abstract, Keywords) & 2017 - 2022         & 15/05/2022 (1:46 pm)  & 5 & 5                    \\ \hline

\end{tabular}
}
\end{table*}
\clearpage
\textbf{Main study}
\begin{table*}[h]
\centering
\label{table:primary_search_string}
\resizebox{\textwidth}{!}{%
\renewcommand{\arraystretch}{1.3}
\begin{tabular}{|p{2.1cm}|P{6cm}|P{2.7cm}|P{1.6cm}|p{1.7cm}|p{1cm}|}

\hline
\textbf{Digital Library}     & \textbf{Search String}                                                                                                                                                                                                                        & \textbf{Search Within}    & \textbf{Time Frame} & \textbf{Search Time}   & \textbf{No. of Papers} \\ \hline
ACM Digital Library & {[}{[}Abstract: ``artificial intelligence''{]} OR {[}Abstract: ``machine learning''{]}{]} AND {[}Abstract: diversity{]} AND {[}{[}Abstract: inclusion{]} OR {[}Abstract: inclusive{]} OR {[}Abstract: inclusiveness{]}{]} & Research article (Abstract)          & 2017 - 2022         & 11/07/2022 (4:23 pm)   & 92                     \\ \hline
IEEE Xplore         & (``Abstract'':``artificial intelligence'' OR ``Abstract'':``machine learning'') AND (``Abstract'':diversity) AND (``Abstract'':inclusion OR ``Abstract'':inclusive OR ``Abstract'':inclusiveness)                                                            & Conference, Journal (Abstract)       & 2017 - 2022         & 07/07/2022 (3:29 am)   & 8                      \\ \hline
Science Direct      & (``artificial intelligence'' OR ``machine learning'') AND ``diversity'' AND (``inclusion'' OR ``inclusive'' OR ``inclusiveness'')                                                                                                                        & Research articles (Title, Abstract, Keywords)         & 2017 - 2022         & 12/07/2022   (5:49 pm) & 12                     \\ \hline
Scopus              & (TITLE-ABS-KEY (``artificial intelligence'' OR ``machine learning'') AND TITLE-ABS-KEY (diversity) AND TITLE-ABS-KEY (inclusion OR inclusive OR inclusiveness))                                                                                  & Conference paper, Article (Title, Abstract, Keywords) & 2017 - 2022         & 12/07/2022   (6:33 pm) & 87                     \\ \hline

\end{tabular}
}
\end{table*}

\clearpage


\section{Mapping of Challenges with Corresponding Solutions about D\&I in AI (RQ1)}
\label{appendix_C}
\textbf{(NoS=No solution, N/A=Not applicable, N=None, G=Gender, S=Sex, A=Age, R=Race, E=Ethnicity, D=Disability, K=Skin tone, L=Geographic location)}

\begin{table*}[h]
\centering
\label{table:challenges_solutions_mapping_DI_AI}
\resizebox{\textwidth}{!}{%
\renewcommand{\arraystretch}{1.3}
\begin{tabular}{|p{3.5cm}|p{3.5cm}|p{3.5cm}|p{3.5cm}|p{3.5cm}|}

\hline
\rowcolor[HTML]{D0CECE} 
\multicolumn{1}{|c|}{\cellcolor[HTML]{D0CECE}\textbf{Paper ID}} & \multicolumn{1}{c|}{\cellcolor[HTML]{D0CECE}\textbf{Challenge ID}} & \multicolumn{1}{c|}{\cellcolor[HTML]{D0CECE}\textbf{Attributes of Challenges}} & \multicolumn{1}{c|}{\cellcolor[HTML]{D0CECE}\textbf{Solution ID}} & \multicolumn{1}{c|}{\cellcolor[HTML]{D0CECE}\textbf{Attributes of Solutions}} \\ \hline
S1                                                              & C1                                                                 & N                                                                              & NoS                                                               & N/A                                                                           \\ \hline
S2                                                              & C2                                                                 & G                                                                              & L3                                                                & G                                                                             \\ \hline
S2                                                              & C3                                                                 & G                                                                              & NoS                                                               & N/A                                                                           \\ \hline
S2                                                              & C4                                                                 & N                                                                              & L2                                                                & N                                                                             \\ \hline
S2                                                              & C5                                                                 & G                                                                              & L1                                                                & N                                                                             \\ \hline
S2                                                              & C6                                                                 & N                                                                              & NoS                                                               & N/A                                                                           \\ \hline
S5                                                              & C7                                                                 & N                                                                              & L4                                                                & N                                                                             \\ \hline
S5                                                              & C8                                                                 & N                                                                              & NoS                                                               & N/A                                                                           \\ \hline
S7                                                              & C9                                                                 & N                                                                              & L5                                                                & R                                                                             \\ \hline
S8                                                              & C7                                                                 & N                                                                              & L6                                                                & N                                                                             \\ \hline
S9                                                              & C10                                                                & N                                                                              & NoS                                                               & N/A                                                                           \\ \hline
S9                                                              & C11                                                                & N                                                                              & NoS                                                               & N/A                                                                           \\ \hline
S9                                                              & C12                                                                & N                                                                              & L7                                                                & N                                                                             \\ \hline
S10                                                             & C13                                                                & A                                                                              & NoS                                                               & N/A                                                                           \\ \hline
S10                                                             & C14                                                                & R, E, S, G                                                                     & L8                                                                & N                                                                             \\ \hline
S10                                                             & C15                                                                & N                                                                              & NoS                                                               & N/A                                                                           \\ \hline
S10                                                             & C16                                                                & N                                                                              & NoS                                                               & N/A                                                                           \\ \hline
S10                                                             & C17                                                                & N                                                                              & NoS                                                               & N/A                                                                           \\ \hline
S10                                                             & C18                                                                & N                                                                              & NoS                                                               & N/A                                                                           \\ \hline
S10                                                             & C6                                                                 & N                                                                              & NoS                                                               & N/A                                                                           \\ \hline
S10                                                             & C7                                                                 & N                                                                              & NoS                                                               & N/A                                                                           \\ \hline
S11                                                             & C19                                                                & N                                                                              & NoS                                                               & N/A                                                                           \\ \hline
S11                                                             & C20                                                                & G                                                                              & L10                                                               & G                                                                             \\ \hline
S11                                                             & C21                                                                & G                                                                              & L9                                                                & G                                                                             \\ \hline
S11                                                             & C22                                                                & N                                                                              & NoS                                                               & N/A                                                                           \\ \hline
S15                                                             & C23                                                                & G                                                                              & NoS                                                               & N/A                                                                           \\ \hline
S15                                                             & C24                                                                & G                                                                              & L12                                                               & N                                                                             \\ \hline
S15                                                             & C24                                                                & G                                                                              & L13                                                               & A, G, R                                                                       \\ \hline
S15                                                             & C25                                                                & N                                                                              & L11                                                               & N                                                                             \\ \hline
S15                                                             & C18                                                                & N                                                                              & NoS                                                               & N/A                                                                           \\ \hline
S16                                                             & C26                                                                & N                                                                              & L14                                                               & N                                                                             \\ \hline
S17                                                             & C27                                                                & N                                                                              & NoS                                                               & N/A                                                                           \\ \hline
S17                                                             & C28                                                                & N                                                                              & NoS                                                               & N/A                                                                           \\ \hline
S17                                                             & C29                                                                & N                                                                              & NoS                                                               & N/A                                                                           \\ \hline
S17                                                             & C30                                                                & N                                                                              & L15                                                               & N                                                                             \\ \hline
S17                                                             & C31                                                                & N                                                                              & L15                                                               & N                                                                             \\ \hline
S17                                                             & C12                                                                & N                                                                              & L16                                                               & R                                                                             \\ \hline
S17                                                             & C18                                                                & N                                                                              & NoS                                                               & N/A                                                                           \\ \hline

\end{tabular}
}
\end{table*}
\begin{table*}[h]
\centering
\resizebox{\textwidth}{!}{%
\renewcommand{\arraystretch}{1.3}
\begin{tabular}{|p{3.5cm}|p{3.5cm}|p{3.5cm}|p{3.5cm}|p{3.5cm}|}

\hline
\rowcolor[HTML]{D0CECE} 
\multicolumn{1}{|c|}{\cellcolor[HTML]{D0CECE}\textbf{Paper ID}} & \multicolumn{1}{c|}{\cellcolor[HTML]{D0CECE}\textbf{Challenge ID}} & \multicolumn{1}{c|}{\cellcolor[HTML]{D0CECE}\textbf{Attributes of Challenges}} & \multicolumn{1}{c|}{\cellcolor[HTML]{D0CECE}\textbf{Solution ID}} & \multicolumn{1}{c|}{\cellcolor[HTML]{D0CECE}\textbf{Attributes of Solutions}} \\ \hline
S18                                                             & C32                                                                & K                                                                              & L17                                                               & K                                                                             \\ \hline
S21                                                             & C33                                                                & N                                                                              & NoS                                                               & N/A                                                                           \\ \hline
S22                                                             & C34                                                                & N                                                                              & L20                                                               & N                                                                             \\ \hline
S22                                                             & C35                                                                & N                                                                              & L18                                                               & N                                                                             \\ \hline
S22                                                             & C35                                                                & N                                                                              & L19                                                               & N                                                                             \\ \hline
S22                                                             & C35                                                                & N                                                                              & L21                                                               & N                                                                             \\ \hline
S23                                                             & C36                                                                & N                                                                              & L22                                                               & R                                                                             \\ \hline
S23                                                             & C36                                                                & N                                                                              & L23                                                               & N                                                                             \\ \hline
S24                                                             & C37                                                                & N                                                                              & L24                                                               & N                                                                             \\ \hline
S24                                                             & C38                                                                & N                                                                              & NoS                                                               & N/A                                                                           \\ \hline
S25                                                             & C17                                                                & N                                                                              & L25                                                               & N                                                                             \\ \hline
S25                                                             & C22                                                                & N                                                                              & L25                                                               & N                                                                             \\ \hline
S25                                                             & C22                                                                & N                                                                              & L26                                                               & N                                                                             \\ \hline
S26                                                             & C39                                                                & R, E                                                                           & NoS                                                               & N/A                                                                           \\ \hline
S27                                                             & C40                                                                & N                                                                              & L8                                                                & N                                                                             \\ \hline
S27                                                             & C41                                                                & G                                                                              & L27                                                               & G                                                                             \\ \hline
S27                                                             & C12                                                                & N                                                                              & L25                                                               & N                                                                             \\ \hline
S29                                                             & C42                                                                & N                                                                              & NoS                                                               & N/A                                                                           \\ \hline
S29                                                             & C43                                                                & G                                                                              & NoS                                                               & N/A                                                                           \\ \hline
S29                                                             & C18                                                                & N                                                                              & NoS                                                               & N/A                                                                           \\ \hline
S30                                                             & C44                                                                & N                                                                              & NoS                                                               & N/A                                                                           \\ \hline
S31                                                             & C45                                                                & G                                                                              & NoS                                                               & N/A                                                                           \\ \hline
S31                                                             & C46                                                                & G                                                                              & L3                                                                & G                                                                             \\ \hline
S33                                                             & C47                                                                & G, R                                                                           & NoS                                                               & N/A                                                                           \\ \hline
S34                                                             & C25                                                                & N                                                                              & NoS                                                               & N/A                                                                           \\ \hline
S35                                                             & C48                                                                & N                                                                              & L28                                                               & L                                                                             \\ \hline
S36                                                             & C11                                                                & N                                                                              & NoS                                                               & N/A                                                                           \\ \hline
S36                                                             & C46                                                                & G                                                                              & NoS                                                               & N/A                                                                           \\ \hline
S37                                                             & C11                                                                & N                                                                              & L8                                                                & N                                                                             \\ \hline
S38                                                             & C49                                                                & D                                                                              & NoS                                                               & N/A                                                                           \\ \hline
S39                                                             & C50                                                                & G                                                                              & L29                                                               & G                                                                             \\ \hline
S39                                                             & C50                                                                & G                                                                              & L30                                                               & G                                                                             \\ \hline
S40                                                             & C51                                                                & G                                                                              & L31                                                               & G                                                                             \\ \hline
S41                                                             & C52                                                                & R                                                                              & L32                                                               & R                                                                             \\ \hline
S42                                                             & C53                                                                & D                                                                              & NoS                                                               & N/A                                                                           \\ \hline
S42                                                             & C17                                                                & N                                                                              & NoS                                                               & N/A                                                                           \\ \hline
S43                                                             & C2                                                                 & G                                                                              & NoS                                                               & N/A                                                                           \\ \hline
S44                                                             & C11                                                                & N                                                                              & L33                                                               & N                                                                             \\ \hline
S46                                                             & C54                                                                & N                                                                              & NoS                                                               & N/A                                                                           \\ \hline
S47                                                             & C55                                                                & N                                                                              & NoS                                                               & N/A                                                                           \\ \hline

\end{tabular}
}
\end{table*}

\clearpage



\section{Mapping of Challenges with Corresponding Solutions about AI for D\&I (RQ2)}
\label{appendix_D}
\textbf{(NoS=No solution, N/A=Not applicable, N=None, G=Gender, S=Sex, A=Age, R=Race, E=Ethnicity, D=Disability, K=Skin tone, L=Geographic location)}

\begin{table*}[h]
\centering
\label{table:challenges_solutions_mapping_RQ2}
\resizebox{\textwidth}{!}{%
\renewcommand{\arraystretch}{1.3}
\begin{tabular}{|p{3.5cm}|p{3.5cm}|p{3.5cm}|p{3.5cm}|p{3.5cm}|}

\hline
\rowcolor[HTML]{D0CECE} 
\multicolumn{1}{|c|}{\cellcolor[HTML]{D0CECE}\textbf{Paper ID}} & \multicolumn{1}{c|}{\cellcolor[HTML]{D0CECE}\textbf{Challenge ID}} & \multicolumn{1}{c|}{\cellcolor[HTML]{D0CECE}\textbf{Attributes of Challenges}} & \multicolumn{1}{c|}{\cellcolor[HTML]{D0CECE}\textbf{Solution ID}} & \multicolumn{1}{c|}{\cellcolor[HTML]{D0CECE}\textbf{Attributes of Solutions}} \\ \hline
S3                                                              & H1                                                                 & R, E                                                                           & N1                                                                & N                                                                             \\ \hline
S4                                                              & H2                                                                 & N                                                                              & N2                                                                & N                                                                             \\ \hline
S4                                                              & H2                                                                 & N                                                                              & N3                                                                & N                                                                             \\ \hline
S4                                                              & H2                                                                 & N                                                                              & N4                                                                & N                                                                             \\ \hline
S6                                                              & \cellcolor[HTML]{FFFFFF}H3                                         & \cellcolor[HTML]{FFFFFF}A, G, R                                                & NoS                                                               & N/A                                                                           \\ \hline
S8                                                              & H4                                                                 & N                                                                              & N5                                                                & N                                                                             \\ \hline
S11                                                             & H5                                                                 & G                                                                              & NoS                                                               & N/A                                                                           \\ \hline
S11                                                             & H6                                                                 & \cellcolor[HTML]{FFFFFF}G                                                      & N6                                                                & G                                                                             \\ \hline
S11                                                             & \cellcolor[HTML]{FFFFFF}H7                                         & \cellcolor[HTML]{FFFFFF}N                                                      & N7                                                                & N                                                                             \\ \hline
S11                                                             & \cellcolor[HTML]{FFFFFF}H7                                         & \cellcolor[HTML]{FFFFFF}N                                                      & N9                                                                & N/A                                                                           \\ \hline
S11                                                             & H8                                                                 & \cellcolor[HTML]{FFFFFF}G                                                      & N8                                                                & G                                                                             \\ \hline
S11                                                             & H8                                                                 & \cellcolor[HTML]{FFFFFF}G                                                      & N7                                                                & N                                                                             \\ \hline
S11                                                             & H8                                                                 & \cellcolor[HTML]{FFFFFF}G                                                      & N9                                                                & G                                                                             \\ \hline
S12                                                             & H9                                                                 & D                                                                              & N10                                                               & N                                                                             \\ \hline
S13                                                             & H6                                                                 & G                                                                              & NoS                                                               & N/A                                                                           \\ \hline
S15                                                             & H10                                                                & N                                                                              & \cellcolor[HTML]{FFFFFF}N11                                       & \cellcolor[HTML]{FFFFFF}N                                                     \\ \hline
S18                                                             & H11                                                                & K                                                                              & \cellcolor[HTML]{FFFFFF}N12                                       & \cellcolor[HTML]{FFFFFF}K                                                     \\ \hline
S19                                                             & H12                                                                & N                                                                              & N13                                                               & N                                                                             \\ \hline
S19                                                             & H12                                                                & N                                                                              & N14                                                               & N                                                                             \\ \hline
S20                                                             & \cellcolor[HTML]{FFFFFF}H13                                        & \cellcolor[HTML]{FFFFFF}N                                                      & NoS                                                               & N/A                                                                           \\ \hline
S21                                                             & H14                                                                & L, G, A                                                                        & N15                                                               & N                                                                             \\ \hline
S22                                                             & H15                                                                & N                                                                              & N16                                                               & N                                                                             \\ \hline
S27                                                             & H16                                                                & G                                                                              & N17                                                               & G                                                                             \\ \hline
S28                                                             & H17                                                                & N                                                                              & N18                                                               & N                                                                             \\ \hline
S32                                                             & \cellcolor[HTML]{FFFFFF}H18                                        & \cellcolor[HTML]{FFFFFF}U                                                      & \cellcolor[HTML]{FFFFFF}N19                                       & \cellcolor[HTML]{FFFFFF}U                                                     \\ \hline
S38                                                             & H19                                                                & D                                                                              & NoS                                                               & N/A                                                                           \\ \hline
S38                                                             & H20                                                                & D                                                                              & N20                                                               & D                                                                             \\ \hline
S38                                                             & H21                                                                & D                                                                              & N20                                                               & D                                                                             \\ \hline
S42                                                             & H22                                                                & D                                                                              & \cellcolor[HTML]{FFFFFF}N21                                       & \cellcolor[HTML]{FFFFFF}D                                                     \\ \hline
S47                                                             & H23                                                                & N                                                                              & N22                                                               & N                                                                             \\ \hline
S48                                                             & H24                                                                & N                                                                              & N23                                                               & N                                                                             \\ \hline

\end{tabular}
}
\end{table*}

\clearpage


\section{Challenges and Solutions with corresponding D\&I pillars for D\&I in AI (RQ1) and AI for D\&I (RQ2)}
\label{appendix_E}
\textbf{(H=Humans, D=Data, P=Process, S=System, G=Governance, O=Other)}

\begin{table*}[h]
\centering
\label{table:pillars}
\resizebox{\textwidth}{!}{%
\renewcommand{\arraystretch}{0.97}
\begin{tabular}{|p{2cm}p{2cm}p{2cm}p{2cm}|p{2cm}p{2cm}p{2cm}p{2cm}|}

\hline
\multicolumn{4}{|c|}{\cellcolor[HTML]{A9D08E}\textbf{D\&I in AI (RQ1)}}                                                                                                                                                                                                     & \multicolumn{4}{c|}{\cellcolor[HTML]{8EA9DB}\textbf{AI for D\&I (RQ2)}}                                                                                                                                                                                                    \\ \hline
\multicolumn{1}{|c|}{\cellcolor[HTML]{C6E0B4}\textbf{Challenge ID}} & \multicolumn{1}{c|}{\cellcolor[HTML]{C6E0B4}\textbf{Pillar}} & \multicolumn{1}{c|}{\cellcolor[HTML]{C6E0B4}\textbf{Solution ID}} & \multicolumn{1}{c|}{\cellcolor[HTML]{C6E0B4}\textbf{Pillar}} & \multicolumn{1}{c|}{\cellcolor[HTML]{B4C6E7}\textbf{Challenge ID}} & \multicolumn{1}{c|}{\cellcolor[HTML]{B4C6E7}\textbf{Pillar}} & \multicolumn{1}{c|}{\cellcolor[HTML]{B4C6E7}\textbf{Solution ID}} & \multicolumn{1}{c|}{\cellcolor[HTML]{B4C6E7}\textbf{Pillar}} \\ \hline
\multicolumn{1}{|l|}{\cellcolor[HTML]{E2EFDA}C1}                    & \multicolumn{1}{l|}{\cellcolor[HTML]{E2EFDA}H, G}            & \multicolumn{1}{l|}{\cellcolor[HTML]{E2EFDA}L1}                   & \multicolumn{1}{l|}{\cellcolor[HTML]{E2EFDA}H, G}                                                         & \multicolumn{1}{l|}{\cellcolor[HTML]{DDEBF7}H1}                    & \multicolumn{1}{l|}{\cellcolor[HTML]{DDEBF7}H, S}            & \multicolumn{1}{l|}{\cellcolor[HTML]{DDEBF7}N1}                   & \multicolumn{1}{l|}{\cellcolor[HTML]{DDEBF7}P, S}                                                         \\ \hline
\multicolumn{1}{|l|}{\cellcolor[HTML]{E2EFDA}C2}                    & \multicolumn{1}{l|}{\cellcolor[HTML]{E2EFDA}H}               & \multicolumn{1}{l|}{\cellcolor[HTML]{E2EFDA}L2}                   & \multicolumn{1}{l|}{\cellcolor[HTML]{E2EFDA}H, G}                                                         & \multicolumn{1}{l|}{\cellcolor[HTML]{DDEBF7}H2}                    & \multicolumn{1}{l|}{\cellcolor[HTML]{DDEBF7}S, G}            & \multicolumn{1}{l|}{\cellcolor[HTML]{DDEBF7}N2}                   & \multicolumn{1}{l|}{\cellcolor[HTML]{DDEBF7}H, S, P}                                                      \\ \hline
\multicolumn{1}{|l|}{\cellcolor[HTML]{E2EFDA}C3}                    & \multicolumn{1}{l|}{\cellcolor[HTML]{E2EFDA}H, G}            & \multicolumn{1}{l|}{\cellcolor[HTML]{E2EFDA}L3}                   & \multicolumn{1}{l|}{\cellcolor[HTML]{E2EFDA}H, G}                                                         & \multicolumn{1}{l|}{\cellcolor[HTML]{DDEBF7}H3}                    & \multicolumn{1}{l|}{\cellcolor[HTML]{DDEBF7}H, D, P, S}      & \multicolumn{1}{l|}{\cellcolor[HTML]{DDEBF7}N3}                   & \multicolumn{1}{l|}{\cellcolor[HTML]{DDEBF7}P, S, G}                                                      \\ \hline
\multicolumn{1}{|l|}{\cellcolor[HTML]{E2EFDA}C4}                    & \multicolumn{1}{l|}{\cellcolor[HTML]{E2EFDA}H, G}            & \multicolumn{1}{l|}{\cellcolor[HTML]{E2EFDA}L4}                   & \multicolumn{1}{l|}{\cellcolor[HTML]{E2EFDA}D, P, S}                                                      & \multicolumn{1}{l|}{\cellcolor[HTML]{DDEBF7}H4}                    & \multicolumn{1}{l|}{\cellcolor[HTML]{DDEBF7}H, G}            & \multicolumn{1}{l|}{\cellcolor[HTML]{DDEBF7}N4}                   & \multicolumn{1}{l|}{\cellcolor[HTML]{DDEBF7}D, P, S}                                                      \\ \hline
\multicolumn{1}{|l|}{\cellcolor[HTML]{E2EFDA}C5}                    & \multicolumn{1}{l|}{\cellcolor[HTML]{E2EFDA}H, G}            & \multicolumn{1}{l|}{\cellcolor[HTML]{E2EFDA}L5}                   & \multicolumn{1}{l|}{\cellcolor[HTML]{E2EFDA}P, S}                                                         & \multicolumn{1}{l|}{\cellcolor[HTML]{DDEBF7}H5}                    & \multicolumn{1}{l|}{\cellcolor[HTML]{DDEBF7}H, D, P, S}      & \multicolumn{1}{l|}{\cellcolor[HTML]{DDEBF7}N5}                   & \multicolumn{1}{l|}{\cellcolor[HTML]{DDEBF7}H, D, S, G}                                                   \\ \hline
\multicolumn{1}{|l|}{\cellcolor[HTML]{E2EFDA}C6}                    & \multicolumn{1}{l|}{\cellcolor[HTML]{E2EFDA}H}               & \multicolumn{1}{l|}{\cellcolor[HTML]{E2EFDA}L6}                   & \multicolumn{1}{l|}{\cellcolor[HTML]{E2EFDA}D}                                                            & \multicolumn{1}{l|}{\cellcolor[HTML]{DDEBF7}H6}                    & \multicolumn{1}{l|}{\cellcolor[HTML]{DDEBF7}H, D, S}         & \multicolumn{1}{l|}{\cellcolor[HTML]{DDEBF7}N6}                   & \multicolumn{1}{l|}{\cellcolor[HTML]{DDEBF7}H, P, S, G}                                                   \\ \hline
\multicolumn{1}{|l|}{\cellcolor[HTML]{E2EFDA}C7}                    & \multicolumn{1}{l|}{\cellcolor[HTML]{E2EFDA}D}               & \multicolumn{1}{l|}{\cellcolor[HTML]{E2EFDA}L7}                   & \multicolumn{1}{l|}{\cellcolor[HTML]{E2EFDA}D}                                                            & \multicolumn{1}{l|}{\cellcolor[HTML]{DDEBF7}H7}                    & \multicolumn{1}{l|}{\cellcolor[HTML]{DDEBF7}D}               & \multicolumn{1}{l|}{\cellcolor[HTML]{DDEBF7}N7}                   & \multicolumn{1}{l|}{\cellcolor[HTML]{DDEBF7}D, P, S}                                                      \\ \hline
\multicolumn{1}{|l|}{\cellcolor[HTML]{E2EFDA}C8}                    & \multicolumn{1}{l|}{\cellcolor[HTML]{E2EFDA}H, G}            & \multicolumn{1}{l|}{\cellcolor[HTML]{E2EFDA}L8}                   & \multicolumn{1}{l|}{\cellcolor[HTML]{E2EFDA}H, S, P}                                                      & \multicolumn{1}{l|}{\cellcolor[HTML]{DDEBF7}H8}                    & \multicolumn{1}{l|}{\cellcolor[HTML]{DDEBF7}H, D}            & \multicolumn{1}{l|}{\cellcolor[HTML]{DDEBF7}N8}                   & \multicolumn{1}{l|}{\cellcolor[HTML]{DDEBF7}D, P, G}                                                      \\ \hline
\multicolumn{1}{|l|}{\cellcolor[HTML]{E2EFDA}C9}                    & \multicolumn{1}{l|}{\cellcolor[HTML]{E2EFDA}P, S}            & \multicolumn{1}{l|}{\cellcolor[HTML]{E2EFDA}L9}                   & \multicolumn{1}{l|}{\cellcolor[HTML]{E2EFDA}S, G}                                                         & \multicolumn{1}{l|}{\cellcolor[HTML]{DDEBF7}H9}                    & \multicolumn{1}{l|}{\cellcolor[HTML]{DDEBF7}H, D, P}         & \multicolumn{1}{l|}{\cellcolor[HTML]{DDEBF7}N9}                   & \multicolumn{1}{l|}{\cellcolor[HTML]{DDEBF7}D, S}                                                         \\ \hline
\multicolumn{1}{|l|}{\cellcolor[HTML]{E2EFDA}C10}                   & \multicolumn{1}{l|}{\cellcolor[HTML]{E2EFDA}H, D}            & \multicolumn{1}{l|}{\cellcolor[HTML]{E2EFDA}L10}                  & \multicolumn{1}{l|}{\cellcolor[HTML]{E2EFDA}H, D, P, G}                                                   & \multicolumn{1}{l|}{\cellcolor[HTML]{DDEBF7}H10}                   & \multicolumn{1}{l|}{\cellcolor[HTML]{DDEBF7}H, P, S, G}      & \multicolumn{1}{l|}{\cellcolor[HTML]{DDEBF7}N10}                  & \multicolumn{1}{l|}{\cellcolor[HTML]{DDEBF7}H, D, P, S}                                                   \\ \hline
\multicolumn{1}{|l|}{\cellcolor[HTML]{E2EFDA}C11}                   & \multicolumn{1}{l|}{\cellcolor[HTML]{E2EFDA}H, P}            & \multicolumn{1}{l|}{\cellcolor[HTML]{E2EFDA}L11}                  & \multicolumn{1}{l|}{\cellcolor[HTML]{E2EFDA}D}                                                            & \multicolumn{1}{l|}{\cellcolor[HTML]{DDEBF7}H11}                   & \multicolumn{1}{l|}{\cellcolor[HTML]{DDEBF7}H, S, P}         & \multicolumn{1}{l|}{\cellcolor[HTML]{DDEBF7}N11}                  & \multicolumn{1}{l|}{\cellcolor[HTML]{DDEBF7}H, D, P, S, G}                                                \\ \hline
\multicolumn{1}{|l|}{\cellcolor[HTML]{E2EFDA}C12}                   & \multicolumn{1}{l|}{\cellcolor[HTML]{E2EFDA}D}               & \multicolumn{1}{l|}{\cellcolor[HTML]{E2EFDA}L12}                  & \multicolumn{1}{l|}{\cellcolor[HTML]{E2EFDA}D, P, S, G}                                                   & \multicolumn{1}{l|}{\cellcolor[HTML]{DDEBF7}H12}                   & \multicolumn{1}{l|}{\cellcolor[HTML]{DDEBF7}S, G}            & \multicolumn{1}{l|}{\cellcolor[HTML]{DDEBF7}N12}                  & \multicolumn{1}{l|}{\cellcolor[HTML]{DDEBF7}D, P, S}                                                      \\ \hline
\multicolumn{1}{|l|}{\cellcolor[HTML]{E2EFDA}C13}                   & \multicolumn{1}{l|}{\cellcolor[HTML]{E2EFDA}H, P}            & \multicolumn{1}{l|}{\cellcolor[HTML]{E2EFDA}L13}                  & \multicolumn{1}{l|}{\cellcolor[HTML]{E2EFDA}H, D, P, G}                                                   & \multicolumn{1}{l|}{\cellcolor[HTML]{DDEBF7}H13}                   & \multicolumn{1}{l|}{\cellcolor[HTML]{DDEBF7}H, G}            & \multicolumn{1}{l|}{\cellcolor[HTML]{DDEBF7}N13}                  & \multicolumn{1}{l|}{\cellcolor[HTML]{DDEBF7}D, P, S, G}                                                   \\ \hline
\multicolumn{1}{|l|}{\cellcolor[HTML]{E2EFDA}C14}                   & \multicolumn{1}{l|}{\cellcolor[HTML]{E2EFDA}H, D, P, S}      & \multicolumn{1}{l|}{\cellcolor[HTML]{E2EFDA}L14}                  & \multicolumn{1}{l|}{\cellcolor[HTML]{E2EFDA}P, S}                                                         & \multicolumn{1}{l|}{\cellcolor[HTML]{DDEBF7}H14}                   & \multicolumn{1}{l|}{\cellcolor[HTML]{DDEBF7}H, G}            & \multicolumn{1}{l|}{\cellcolor[HTML]{DDEBF7}N14}                  & \multicolumn{1}{l|}{\cellcolor[HTML]{DDEBF7}D, P, S, G}                                                   \\ \hline
\multicolumn{1}{|l|}{\cellcolor[HTML]{E2EFDA}C15}                   & \multicolumn{1}{l|}{\cellcolor[HTML]{E2EFDA}H, P, G}         & \multicolumn{1}{l|}{\cellcolor[HTML]{E2EFDA}L15}                  & \multicolumn{1}{l|}{\cellcolor[HTML]{E2EFDA}D, G}                                                         & \multicolumn{1}{l|}{\cellcolor[HTML]{DDEBF7}H15}                   & \multicolumn{1}{l|}{\cellcolor[HTML]{DDEBF7}P, S}            & \multicolumn{1}{l|}{\cellcolor[HTML]{DDEBF7}N15}                  & \multicolumn{1}{l|}{\cellcolor[HTML]{DDEBF7}D, P, S}                                                      \\ \hline
\multicolumn{1}{|l|}{\cellcolor[HTML]{E2EFDA}C16}                   & \multicolumn{1}{l|}{\cellcolor[HTML]{E2EFDA}H, P, G}         & \multicolumn{1}{l|}{\cellcolor[HTML]{E2EFDA}L16}                  & \multicolumn{1}{l|}{\cellcolor[HTML]{E2EFDA}D, P}                                                         & \multicolumn{1}{l|}{\cellcolor[HTML]{DDEBF7}H16}                   & \multicolumn{1}{l|}{\cellcolor[HTML]{DDEBF7}D, P, S}         & \multicolumn{1}{l|}{\cellcolor[HTML]{DDEBF7}N16}                  & \multicolumn{1}{l|}{\cellcolor[HTML]{DDEBF7}P, S}                                                         \\ \hline
\multicolumn{1}{|l|}{\cellcolor[HTML]{E2EFDA}C17}                   & \multicolumn{1}{l|}{\cellcolor[HTML]{E2EFDA}H, D, P, S}      & \multicolumn{1}{l|}{\cellcolor[HTML]{E2EFDA}L17}                  & \multicolumn{1}{l|}{\cellcolor[HTML]{E2EFDA}D, P}                                                         & \multicolumn{1}{l|}{\cellcolor[HTML]{DDEBF7}H17}                   & \multicolumn{1}{l|}{\cellcolor[HTML]{DDEBF7}P, S}            & \multicolumn{1}{l|}{\cellcolor[HTML]{DDEBF7}N17}                  & \multicolumn{1}{l|}{\cellcolor[HTML]{DDEBF7}H, D, P, S}                                                   \\ \hline
\multicolumn{1}{|l|}{\cellcolor[HTML]{E2EFDA}C18}                   & \multicolumn{1}{l|}{\cellcolor[HTML]{E2EFDA}H, D, P, S}      & \multicolumn{1}{l|}{\cellcolor[HTML]{E2EFDA}L18}                  & \multicolumn{1}{l|}{\cellcolor[HTML]{E2EFDA}H, D, P}                                                      & \multicolumn{1}{l|}{\cellcolor[HTML]{DDEBF7}H18}                   & \multicolumn{1}{l|}{\cellcolor[HTML]{DDEBF7}H, S, P}         & \multicolumn{1}{l|}{\cellcolor[HTML]{DDEBF7}N18}                  & \multicolumn{1}{l|}{\cellcolor[HTML]{DDEBF7}P, S}                                                         \\ \hline
\multicolumn{1}{|l|}{\cellcolor[HTML]{E2EFDA}C19}                   & \multicolumn{1}{l|}{\cellcolor[HTML]{E2EFDA}D}               & \multicolumn{1}{l|}{\cellcolor[HTML]{E2EFDA}L19}                  & \multicolumn{1}{l|}{\cellcolor[HTML]{E2EFDA}H, S, P}                                                      & \multicolumn{1}{l|}{\cellcolor[HTML]{DDEBF7}H19}                   & \multicolumn{1}{l|}{\cellcolor[HTML]{DDEBF7}D, P, S}         & \multicolumn{1}{l|}{\cellcolor[HTML]{DDEBF7}N19}                  & \multicolumn{1}{l|}{\cellcolor[HTML]{DDEBF7}H, S, P}                                                      \\ \hline
\multicolumn{1}{|l|}{\cellcolor[HTML]{E2EFDA}C20}                   & \multicolumn{1}{l|}{\cellcolor[HTML]{E2EFDA}H, D, G}         & \multicolumn{1}{l|}{\cellcolor[HTML]{E2EFDA}L20}                  & \multicolumn{1}{l|}{\cellcolor[HTML]{E2EFDA}P, S}                                                         & \multicolumn{1}{l|}{\cellcolor[HTML]{DDEBF7}H20}                   & \multicolumn{1}{l|}{\cellcolor[HTML]{DDEBF7}H, S, G}         & \multicolumn{1}{l|}{\cellcolor[HTML]{DDEBF7}N20}                  & \multicolumn{1}{l|}{\cellcolor[HTML]{DDEBF7}P, S, G}                                                      \\ \hline
\multicolumn{1}{|l|}{\cellcolor[HTML]{E2EFDA}C21}                   & \multicolumn{1}{l|}{\cellcolor[HTML]{E2EFDA}S}               & \multicolumn{1}{l|}{\cellcolor[HTML]{E2EFDA}L21}                  & \multicolumn{1}{l|}{\cellcolor[HTML]{E2EFDA}H, G}                                                         & \multicolumn{1}{l|}{\cellcolor[HTML]{DDEBF7}H21}                   & \multicolumn{1}{l|}{\cellcolor[HTML]{DDEBF7}H, S, G}         & \multicolumn{1}{l|}{\cellcolor[HTML]{DDEBF7}N21}                  & \multicolumn{1}{l|}{\cellcolor[HTML]{DDEBF7}S, G}                                                         \\ \hline
\multicolumn{1}{|l|}{\cellcolor[HTML]{E2EFDA}C22}                   & \multicolumn{1}{l|}{\cellcolor[HTML]{E2EFDA}D, P, S}         & \multicolumn{1}{l|}{\cellcolor[HTML]{E2EFDA}L22}                  & \multicolumn{1}{l|}{\cellcolor[HTML]{E2EFDA}H, G}                                                         & \multicolumn{1}{l|}{\cellcolor[HTML]{DDEBF7}H22}                   & \multicolumn{1}{l|}{\cellcolor[HTML]{DDEBF7}H, S, G}         & \multicolumn{1}{l|}{\cellcolor[HTML]{DDEBF7}N22}                  & \multicolumn{1}{l|}{\cellcolor[HTML]{DDEBF7}P, S}                                                         \\ \hline
\multicolumn{1}{|l|}{\cellcolor[HTML]{E2EFDA}C23}                   & \multicolumn{1}{l|}{\cellcolor[HTML]{E2EFDA}H}               & \multicolumn{1}{l|}{\cellcolor[HTML]{E2EFDA}L23}                  & \multicolumn{1}{l|}{\cellcolor[HTML]{E2EFDA}H, P, G}                                                      & \multicolumn{1}{l|}{\cellcolor[HTML]{DDEBF7}H23}                   & \multicolumn{1}{l|}{\cellcolor[HTML]{DDEBF7}D, P}            & \multicolumn{1}{l|}{\cellcolor[HTML]{DDEBF7}N23}                  & \multicolumn{1}{l|}{\cellcolor[HTML]{DDEBF7}H, D, S}                                                      \\ \hline
\multicolumn{1}{|l|}{\cellcolor[HTML]{E2EFDA}C24}                   & \multicolumn{1}{l|}{\cellcolor[HTML]{E2EFDA}H, G}            & \multicolumn{1}{l|}{\cellcolor[HTML]{E2EFDA}L24}                  & \multicolumn{1}{l|}{\cellcolor[HTML]{E2EFDA}H, S, G}                                                      & \multicolumn{1}{l|}{\cellcolor[HTML]{DDEBF7}H24}                   & \multicolumn{1}{l|}{\cellcolor[HTML]{DDEBF7}H, S, P}         & \multicolumn{1}{l|}{\cellcolor[HTML]{DDEBF7}-}                    & \multicolumn{1}{l|}{\cellcolor[HTML]{DDEBF7}-}                                                            \\ \hline
\multicolumn{1}{|l|}{\cellcolor[HTML]{E2EFDA}C25}                   & \multicolumn{1}{l|}{\cellcolor[HTML]{E2EFDA}D, P, S}         & \multicolumn{1}{l|}{\cellcolor[HTML]{E2EFDA}L25}                  & \multicolumn{1}{l|}{\cellcolor[HTML]{E2EFDA}H, D, P, S}                                                   & \multicolumn{1}{l|}{\cellcolor[HTML]{DDEBF7}-}                     & \multicolumn{1}{l|}{\cellcolor[HTML]{DDEBF7}-}               & \multicolumn{1}{l|}{\cellcolor[HTML]{DDEBF7}-}                    & \multicolumn{1}{l|}{\cellcolor[HTML]{DDEBF7}-}                                                            \\ \hline
\multicolumn{1}{|l|}{\cellcolor[HTML]{E2EFDA}C26}                   & \multicolumn{1}{l|}{\cellcolor[HTML]{E2EFDA}H, P, G}         & \multicolumn{1}{l|}{\cellcolor[HTML]{E2EFDA}L26}                  & \multicolumn{1}{l|}{\cellcolor[HTML]{E2EFDA}H, D, P, S, G}                                                & \multicolumn{1}{l|}{\cellcolor[HTML]{DDEBF7}-}                     & \multicolumn{1}{l|}{\cellcolor[HTML]{DDEBF7}-}               & \multicolumn{1}{l|}{\cellcolor[HTML]{DDEBF7}-}                    & \multicolumn{1}{l|}{\cellcolor[HTML]{DDEBF7}-}                                                            \\ \hline
\multicolumn{1}{|l|}{\cellcolor[HTML]{E2EFDA}C27}                   & \multicolumn{1}{l|}{\cellcolor[HTML]{E2EFDA}H, P}            & \multicolumn{1}{l|}{\cellcolor[HTML]{E2EFDA}L27}                  & \multicolumn{1}{l|}{\cellcolor[HTML]{E2EFDA}H, D}                                                         & \multicolumn{1}{l|}{\cellcolor[HTML]{DDEBF7}-}                     & \multicolumn{1}{l|}{\cellcolor[HTML]{DDEBF7}-}               & \multicolumn{1}{l|}{\cellcolor[HTML]{DDEBF7}-}                    & \multicolumn{1}{l|}{\cellcolor[HTML]{DDEBF7}-}                                                            \\ \hline
\multicolumn{1}{|l|}{\cellcolor[HTML]{E2EFDA}C28}                   & \multicolumn{1}{l|}{\cellcolor[HTML]{E2EFDA}H, D}            & \multicolumn{1}{l|}{\cellcolor[HTML]{E2EFDA}L28}                  & \multicolumn{1}{l|}{\cellcolor[HTML]{E2EFDA}P, S}                                                         & \multicolumn{1}{l|}{\cellcolor[HTML]{DDEBF7}-}                     & \multicolumn{1}{l|}{\cellcolor[HTML]{DDEBF7}-}               & \multicolumn{1}{l|}{\cellcolor[HTML]{DDEBF7}-}                    & \multicolumn{1}{l|}{\cellcolor[HTML]{DDEBF7}-}                                                            \\ \hline
\multicolumn{1}{|l|}{\cellcolor[HTML]{E2EFDA}C29}                   & \multicolumn{1}{l|}{\cellcolor[HTML]{E2EFDA}D}               & \multicolumn{1}{l|}{\cellcolor[HTML]{E2EFDA}L29}                  & \multicolumn{1}{l|}{\cellcolor[HTML]{E2EFDA}H, D, P}                                                      & \multicolumn{1}{l|}{\cellcolor[HTML]{DDEBF7}-}                     & \multicolumn{1}{l|}{\cellcolor[HTML]{DDEBF7}-}               & \multicolumn{1}{l|}{\cellcolor[HTML]{DDEBF7}-}                    & \multicolumn{1}{l|}{\cellcolor[HTML]{DDEBF7}-}                                                            \\ \hline
\multicolumn{1}{|l|}{\cellcolor[HTML]{E2EFDA}C30}                   & \multicolumn{1}{l|}{\cellcolor[HTML]{E2EFDA}H, S, G}         & \multicolumn{1}{l|}{\cellcolor[HTML]{E2EFDA}L30}                  & \multicolumn{1}{l|}{\cellcolor[HTML]{E2EFDA}H, P}                                                         & \multicolumn{1}{l|}{\cellcolor[HTML]{DDEBF7}-}                     & \multicolumn{1}{l|}{\cellcolor[HTML]{DDEBF7}-}               & \multicolumn{1}{l|}{\cellcolor[HTML]{DDEBF7}-}                    & \multicolumn{1}{l|}{\cellcolor[HTML]{DDEBF7}-}                                                            \\ \hline
\multicolumn{1}{|l|}{\cellcolor[HTML]{E2EFDA}C31}                   & \multicolumn{1}{l|}{\cellcolor[HTML]{E2EFDA}D, P}            & \multicolumn{1}{l|}{\cellcolor[HTML]{E2EFDA}L31}                  & \multicolumn{1}{l|}{\cellcolor[HTML]{E2EFDA}H, D, P, G}                                                   & \multicolumn{1}{l|}{\cellcolor[HTML]{DDEBF7}-}                     & \multicolumn{1}{l|}{\cellcolor[HTML]{DDEBF7}-}               & \multicolumn{1}{l|}{\cellcolor[HTML]{DDEBF7}-}                    & \multicolumn{1}{l|}{\cellcolor[HTML]{DDEBF7}-}                                                            \\ \hline
\multicolumn{1}{|l|}{\cellcolor[HTML]{E2EFDA}C32}                   & \multicolumn{1}{l|}{\cellcolor[HTML]{E2EFDA}H, D, S}         & \multicolumn{1}{l|}{\cellcolor[HTML]{E2EFDA}L32}                  & \multicolumn{1}{l|}{\cellcolor[HTML]{E2EFDA}D, P}                                                         & \multicolumn{1}{l|}{\cellcolor[HTML]{DDEBF7}-}                     & \multicolumn{1}{l|}{\cellcolor[HTML]{DDEBF7}-}               & \multicolumn{1}{l|}{\cellcolor[HTML]{DDEBF7}-}                    & \multicolumn{1}{l|}{\cellcolor[HTML]{DDEBF7}-}                                                            \\ \hline
\multicolumn{1}{|l|}{\cellcolor[HTML]{E2EFDA}C33}                   & \multicolumn{1}{l|}{\cellcolor[HTML]{E2EFDA}D, S}            & \multicolumn{1}{l|}{\cellcolor[HTML]{E2EFDA}L33}                  & \multicolumn{1}{l|}{\cellcolor[HTML]{E2EFDA}P, S}                                                         & \multicolumn{1}{l|}{\cellcolor[HTML]{DDEBF7}-}                     & \multicolumn{1}{l|}{\cellcolor[HTML]{DDEBF7}-}               & \multicolumn{1}{l|}{\cellcolor[HTML]{DDEBF7}-}                    & \multicolumn{1}{l|}{\cellcolor[HTML]{DDEBF7}-}                                                            \\ \hline
\multicolumn{1}{|l|}{\cellcolor[HTML]{E2EFDA}C34}                   & \multicolumn{1}{l|}{\cellcolor[HTML]{E2EFDA}P, S}            & \multicolumn{1}{l|}{\cellcolor[HTML]{E2EFDA}-}                    & \multicolumn{1}{l|}{\cellcolor[HTML]{E2EFDA}-}                                                            & \multicolumn{1}{l|}{\cellcolor[HTML]{DDEBF7}-}                     & \multicolumn{1}{l|}{\cellcolor[HTML]{DDEBF7}-}               & \multicolumn{1}{l|}{\cellcolor[HTML]{DDEBF7}-}                    & \multicolumn{1}{l|}{\cellcolor[HTML]{DDEBF7}-}                                                            \\ \hline
\multicolumn{1}{|l|}{\cellcolor[HTML]{E2EFDA}C35}                   & \multicolumn{1}{l|}{\cellcolor[HTML]{E2EFDA}H, P}            & \multicolumn{1}{l|}{\cellcolor[HTML]{E2EFDA}-}                    & \multicolumn{1}{l|}{\cellcolor[HTML]{E2EFDA}-}                                                            & \multicolumn{1}{l|}{\cellcolor[HTML]{DDEBF7}-}                     & \multicolumn{1}{l|}{\cellcolor[HTML]{DDEBF7}-}               & \multicolumn{1}{l|}{\cellcolor[HTML]{DDEBF7}-}                    & \multicolumn{1}{l|}{\cellcolor[HTML]{DDEBF7}-}                                                            \\ \hline
\multicolumn{1}{|l|}{\cellcolor[HTML]{E2EFDA}C36}                   & \multicolumn{1}{l|}{\cellcolor[HTML]{E2EFDA}P, S, G}         & \multicolumn{1}{l|}{\cellcolor[HTML]{E2EFDA}-}                    & \multicolumn{1}{l|}{\cellcolor[HTML]{E2EFDA}-}                                                            & \multicolumn{1}{l|}{\cellcolor[HTML]{DDEBF7}-}                     & \multicolumn{1}{l|}{\cellcolor[HTML]{DDEBF7}-}               & \multicolumn{1}{l|}{\cellcolor[HTML]{DDEBF7}-}                    & \multicolumn{1}{l|}{\cellcolor[HTML]{DDEBF7}-}                                                            \\ \hline
\multicolumn{1}{|l|}{\cellcolor[HTML]{E2EFDA}C37}                   & \multicolumn{1}{l|}{\cellcolor[HTML]{E2EFDA}H}               & \multicolumn{1}{l|}{\cellcolor[HTML]{E2EFDA}-}                    & \multicolumn{1}{l|}{\cellcolor[HTML]{E2EFDA}-}                                                            & \multicolumn{1}{l|}{\cellcolor[HTML]{DDEBF7}-}                     & \multicolumn{1}{l|}{\cellcolor[HTML]{DDEBF7}-}               & \multicolumn{1}{l|}{\cellcolor[HTML]{DDEBF7}-}                    & \multicolumn{1}{l|}{\cellcolor[HTML]{DDEBF7}-}                                                            \\ \hline
\multicolumn{1}{|l|}{\cellcolor[HTML]{E2EFDA}C38}                   & \multicolumn{1}{l|}{\cellcolor[HTML]{E2EFDA}H, P, S}         & \multicolumn{1}{l|}{\cellcolor[HTML]{E2EFDA}-}                    & \multicolumn{1}{l|}{\cellcolor[HTML]{E2EFDA}-}                                                            & \multicolumn{1}{l|}{\cellcolor[HTML]{DDEBF7}-}                     & \multicolumn{1}{l|}{\cellcolor[HTML]{DDEBF7}-}               & \multicolumn{1}{l|}{\cellcolor[HTML]{DDEBF7}-}                    & \multicolumn{1}{l|}{\cellcolor[HTML]{DDEBF7}-}                                                            \\ \hline
\multicolumn{1}{|l|}{\cellcolor[HTML]{E2EFDA}C39}                   & \multicolumn{1}{l|}{\cellcolor[HTML]{E2EFDA}D}               & \multicolumn{1}{l|}{\cellcolor[HTML]{E2EFDA}-}                    & \multicolumn{1}{l|}{\cellcolor[HTML]{E2EFDA}-}                                                            & \multicolumn{1}{l|}{\cellcolor[HTML]{DDEBF7}-}                     & \multicolumn{1}{l|}{\cellcolor[HTML]{DDEBF7}-}               & \multicolumn{1}{l|}{\cellcolor[HTML]{DDEBF7}-}                    & \multicolumn{1}{l|}{\cellcolor[HTML]{DDEBF7}-}                                                            \\ \hline
\multicolumn{1}{|l|}{\cellcolor[HTML]{E2EFDA}C40}                   & \multicolumn{1}{l|}{\cellcolor[HTML]{E2EFDA}D, P, S}         & \multicolumn{1}{l|}{\cellcolor[HTML]{E2EFDA}-}                    & \multicolumn{1}{l|}{\cellcolor[HTML]{E2EFDA}-}                                                            & \multicolumn{1}{l|}{\cellcolor[HTML]{DDEBF7}-}                     & \multicolumn{1}{l|}{\cellcolor[HTML]{DDEBF7}-}               & \multicolumn{1}{l|}{\cellcolor[HTML]{DDEBF7}-}                    & \multicolumn{1}{l|}{\cellcolor[HTML]{DDEBF7}-}                                                            \\ \hline
\multicolumn{1}{|l|}{\cellcolor[HTML]{E2EFDA}C41}                   & \multicolumn{1}{l|}{\cellcolor[HTML]{E2EFDA}H, G}            & \multicolumn{1}{l|}{\cellcolor[HTML]{E2EFDA}-}                    & \multicolumn{1}{l|}{\cellcolor[HTML]{E2EFDA}-}                                                            & \multicolumn{1}{l|}{\cellcolor[HTML]{DDEBF7}-}                     & \multicolumn{1}{l|}{\cellcolor[HTML]{DDEBF7}-}               & \multicolumn{1}{l|}{\cellcolor[HTML]{DDEBF7}-}                    & \multicolumn{1}{l|}{\cellcolor[HTML]{DDEBF7}-}                                                            \\ \hline
\multicolumn{1}{|l|}{\cellcolor[HTML]{E2EFDA}C42}                   & \multicolumn{1}{l|}{\cellcolor[HTML]{E2EFDA}H, D, P, S}      & \multicolumn{1}{l|}{\cellcolor[HTML]{E2EFDA}-}                    & \multicolumn{1}{l|}{\cellcolor[HTML]{E2EFDA}-}                                                            & \multicolumn{1}{l|}{\cellcolor[HTML]{DDEBF7}-}                     & \multicolumn{1}{l|}{\cellcolor[HTML]{DDEBF7}-}               & \multicolumn{1}{l|}{\cellcolor[HTML]{DDEBF7}-}                    & \multicolumn{1}{l|}{\cellcolor[HTML]{DDEBF7}-}                                                            \\ \hline
\multicolumn{1}{|l|}{\cellcolor[HTML]{E2EFDA}C43}                   & \multicolumn{1}{l|}{\cellcolor[HTML]{E2EFDA}D, P, S}         & \multicolumn{1}{l|}{\cellcolor[HTML]{E2EFDA}-}                    & \multicolumn{1}{l|}{\cellcolor[HTML]{E2EFDA}-}                                                            & \multicolumn{1}{l|}{\cellcolor[HTML]{DDEBF7}-}                     & \multicolumn{1}{l|}{\cellcolor[HTML]{DDEBF7}-}               & \multicolumn{1}{l|}{\cellcolor[HTML]{DDEBF7}-}                    & \multicolumn{1}{l|}{\cellcolor[HTML]{DDEBF7}-}                                                            \\ \hline
\multicolumn{1}{|l|}{\cellcolor[HTML]{E2EFDA}C44}                   & \multicolumn{1}{l|}{\cellcolor[HTML]{E2EFDA}H, D, P}         & \multicolumn{1}{l|}{\cellcolor[HTML]{E2EFDA}-}                    & \multicolumn{1}{l|}{\cellcolor[HTML]{E2EFDA}-}                                                            & \multicolumn{1}{l|}{\cellcolor[HTML]{DDEBF7}-}                     & \multicolumn{1}{l|}{\cellcolor[HTML]{DDEBF7}-}               & \multicolumn{1}{l|}{\cellcolor[HTML]{DDEBF7}-}                    & \multicolumn{1}{l|}{\cellcolor[HTML]{DDEBF7}-}                                                            \\ \hline
\multicolumn{1}{|l|}{\cellcolor[HTML]{E2EFDA}C45}                   & \multicolumn{1}{l|}{\cellcolor[HTML]{E2EFDA}D}               & \multicolumn{1}{l|}{\cellcolor[HTML]{E2EFDA}-}                    & \multicolumn{1}{l|}{\cellcolor[HTML]{E2EFDA}-}                                                            & \multicolumn{1}{l|}{\cellcolor[HTML]{DDEBF7}-}                     & \multicolumn{1}{l|}{\cellcolor[HTML]{DDEBF7}-}               & \multicolumn{1}{l|}{\cellcolor[HTML]{DDEBF7}-}                    & \multicolumn{1}{l|}{\cellcolor[HTML]{DDEBF7}-}                                                            \\ \hline
\multicolumn{1}{|l|}{\cellcolor[HTML]{E2EFDA}C46}                   & \multicolumn{1}{l|}{\cellcolor[HTML]{E2EFDA}H, S, P}         & \multicolumn{1}{l|}{\cellcolor[HTML]{E2EFDA}-}                    & \multicolumn{1}{l|}{\cellcolor[HTML]{E2EFDA}-}                                                            & \multicolumn{1}{l|}{\cellcolor[HTML]{DDEBF7}-}                     & \multicolumn{1}{l|}{\cellcolor[HTML]{DDEBF7}-}               & \multicolumn{1}{l|}{\cellcolor[HTML]{DDEBF7}-}                    & \multicolumn{1}{l|}{\cellcolor[HTML]{DDEBF7}-}                                                            \\ \hline
\multicolumn{1}{|l|}{\cellcolor[HTML]{E2EFDA}C47}                   & \multicolumn{1}{l|}{\cellcolor[HTML]{E2EFDA}H}               & \multicolumn{1}{l|}{\cellcolor[HTML]{E2EFDA}-}                    & \multicolumn{1}{l|}{\cellcolor[HTML]{E2EFDA}-}                                                            & \multicolumn{1}{l|}{\cellcolor[HTML]{DDEBF7}-}                     & \multicolumn{1}{l|}{\cellcolor[HTML]{DDEBF7}-}               & \multicolumn{1}{l|}{\cellcolor[HTML]{DDEBF7}-}                    & \multicolumn{1}{l|}{\cellcolor[HTML]{DDEBF7}-}                                                            \\ \hline
\multicolumn{1}{|l|}{\cellcolor[HTML]{E2EFDA}C48}                   & \multicolumn{1}{l|}{\cellcolor[HTML]{E2EFDA}H, D, S}         & \multicolumn{1}{l|}{\cellcolor[HTML]{E2EFDA}-}                    & \multicolumn{1}{l|}{\cellcolor[HTML]{E2EFDA}-}                                                            & \multicolumn{1}{l|}{\cellcolor[HTML]{DDEBF7}-}                     & \multicolumn{1}{l|}{\cellcolor[HTML]{DDEBF7}-}               & \multicolumn{1}{l|}{\cellcolor[HTML]{DDEBF7}-}                    & \multicolumn{1}{l|}{\cellcolor[HTML]{DDEBF7}-}                                                            \\ \hline
\multicolumn{1}{|l|}{\cellcolor[HTML]{E2EFDA}C49}                   & \multicolumn{1}{l|}{\cellcolor[HTML]{E2EFDA}P, S, G}         & \multicolumn{1}{l|}{\cellcolor[HTML]{E2EFDA}-}                    & \multicolumn{1}{l|}{\cellcolor[HTML]{E2EFDA}-}                                                            & \multicolumn{1}{l|}{\cellcolor[HTML]{DDEBF7}-}                     & \multicolumn{1}{l|}{\cellcolor[HTML]{DDEBF7}-}               & \multicolumn{1}{l|}{\cellcolor[HTML]{DDEBF7}-}                    & \multicolumn{1}{l|}{\cellcolor[HTML]{DDEBF7}-}                                                            \\ \hline
\multicolumn{1}{|l|}{\cellcolor[HTML]{E2EFDA}C50}                   & \multicolumn{1}{l|}{\cellcolor[HTML]{E2EFDA}P, S}            & \multicolumn{1}{l|}{\cellcolor[HTML]{E2EFDA}-}                    & \multicolumn{1}{l|}{\cellcolor[HTML]{E2EFDA}-}                                                            & \multicolumn{1}{l|}{\cellcolor[HTML]{DDEBF7}-}                     & \multicolumn{1}{l|}{\cellcolor[HTML]{DDEBF7}-}               & \multicolumn{1}{l|}{\cellcolor[HTML]{DDEBF7}-}                    & \multicolumn{1}{l|}{\cellcolor[HTML]{DDEBF7}-}                                                            \\ \hline
\multicolumn{1}{|l|}{\cellcolor[HTML]{E2EFDA}C51}                   & \multicolumn{1}{l|}{\cellcolor[HTML]{E2EFDA}H, P, G}         & \multicolumn{1}{l|}{\cellcolor[HTML]{E2EFDA}-}                    & \multicolumn{1}{l|}{\cellcolor[HTML]{E2EFDA}-}                                                            & \multicolumn{1}{l|}{\cellcolor[HTML]{DDEBF7}-}                     & \multicolumn{1}{l|}{\cellcolor[HTML]{DDEBF7}-}               & \multicolumn{1}{l|}{\cellcolor[HTML]{DDEBF7}-}                    & \multicolumn{1}{l|}{\cellcolor[HTML]{DDEBF7}-}                                                            \\ \hline
\multicolumn{1}{|l|}{\cellcolor[HTML]{E2EFDA}C52}                   & \multicolumn{1}{l|}{\cellcolor[HTML]{E2EFDA}D, P, S}         & \multicolumn{1}{l|}{\cellcolor[HTML]{E2EFDA}-}                    & \multicolumn{1}{l|}{\cellcolor[HTML]{E2EFDA}-}                                                            & \multicolumn{1}{l|}{\cellcolor[HTML]{DDEBF7}-}                     & \multicolumn{1}{l|}{\cellcolor[HTML]{DDEBF7}-}               & \multicolumn{1}{l|}{\cellcolor[HTML]{DDEBF7}-}                    & \multicolumn{1}{l|}{\cellcolor[HTML]{DDEBF7}-}                                                            \\ \hline
\multicolumn{1}{|l|}{\cellcolor[HTML]{E2EFDA}C53}                   & \multicolumn{1}{l|}{\cellcolor[HTML]{E2EFDA}H, D, S}         & \multicolumn{1}{l|}{\cellcolor[HTML]{E2EFDA}-}                    & \multicolumn{1}{l|}{\cellcolor[HTML]{E2EFDA}-}                                                            & \multicolumn{1}{l|}{\cellcolor[HTML]{DDEBF7}-}                     & \multicolumn{1}{l|}{\cellcolor[HTML]{DDEBF7}-}               & \multicolumn{1}{l|}{\cellcolor[HTML]{DDEBF7}-}                    & \multicolumn{1}{l|}{\cellcolor[HTML]{DDEBF7}-}                                                            \\ \hline
\multicolumn{1}{|l|}{\cellcolor[HTML]{E2EFDA}C54}                   & \multicolumn{1}{l|}{\cellcolor[HTML]{E2EFDA}H, D, S}         & \multicolumn{1}{l|}{\cellcolor[HTML]{E2EFDA}-}                    & \multicolumn{1}{l|}{\cellcolor[HTML]{E2EFDA}-}                                                            & \multicolumn{1}{l|}{\cellcolor[HTML]{DDEBF7}-}                     & \multicolumn{1}{l|}{\cellcolor[HTML]{DDEBF7}-}               & \multicolumn{1}{l|}{\cellcolor[HTML]{DDEBF7}-}                    & \multicolumn{1}{l|}{\cellcolor[HTML]{DDEBF7}-}                                                            \\ \hline
\multicolumn{1}{|l|}{\cellcolor[HTML]{E2EFDA}C55}                   & \multicolumn{1}{l|}{\cellcolor[HTML]{E2EFDA}P, S}            & \multicolumn{1}{l|}{\cellcolor[HTML]{E2EFDA}-}                    & \multicolumn{1}{l|}{\cellcolor[HTML]{E2EFDA}-}                                                            & \multicolumn{1}{l|}{\cellcolor[HTML]{DDEBF7}-}                     & \multicolumn{1}{l|}{\cellcolor[HTML]{DDEBF7}-}               & \multicolumn{1}{l|}{\cellcolor[HTML]{DDEBF7}-}                    & \multicolumn{1}{l|}{\cellcolor[HTML]{DDEBF7}-}                                                            \\ \hline

\end{tabular}
}
\end{table*}


\end{appendices}

\end{document}